\pdfoutput=1
\documentclass[acmtog]{acmart}

\makeatletter
\patchcmd{\@mkbibcitation}{\ref{TotPages}~\@pages@word}{12~pages}{}{}
\patchcmd{\@mkbibcitation}{\ref{TotPages}~\@pages@word}{12~pages}{}{}
\makeatother
\renewcommand\footnotetextcopyrightpermission[1]{}
\makeatletter
\AtBeginDocument{%
  \g@addto@macro\ps@firstpagestyle{\fancyfoot[RO,LE]{}}%
  \g@addto@macro\ps@standardpagestyle{\fancyfoot[RO,LE]{}}%
  \pagestyle{standardpagestyle}}
\makeatother

\usepackage[capitalize]{cleveref}
\crefname{table}{Tab.}{Tabs.}
\Crefname{table}{Tab.}{Tabs.}

\usepackage{enumitem}    %
\usepackage{subcaption}  %
\usepackage{stfloats}    %

\usepackage{listings}
\usepackage{algorithm}
\usepackage{algpseudocode}
\usepackage{bibunits}

\newcommand{\themodel}{UniMate}
\newcommand{\thedata}{UniML3D}

\newcommand{\projectpage}{%
  \textcolor{ACMDarkBlue}{\url{https://linzhanmou.com/unimate/}}}

\newcommand{\codeavailability}{%
  Our project page is available at \projectpage.%
}

\newcommand{\appendixref}[1]{Appendix~\labelcref{#1}}
\newcommand{\Appendixref}[1]{Appendix~\labelcref{#1}}

\newcommand{\boldstartspace}[1]{\par \noindent \textbf{#1}\ }

\copyrightyear{2026}
\acmYear{2026}
\setcopyright{cc}
\setcctype{by}
\acmConference[SA Conference Papers '26]{SIGGRAPH Asia 2026 Conference Papers}{December 01--04, 2026}{Kuala Lumpur, Malaysia}
\acmBooktitle{SIGGRAPH Asia 2026 Conference Papers (SA Conference Papers '26), December 01--04, 2026, Kuala Lumpur, Malaysia}
\acmDOI{10.1145/3829340.3842216}
\acmISBN{979-8-4007-2842-6/2026/12}

\graphicspath{{images/}}

\begin{document}
\title{UniMate: One Unified Model to Animate Diverse Skeletons}

\author{Linzhan Mou}
\affiliation{%
  \institution{Princeton University}
  \country{USA}
}
\email{linzhan@princeton.edu}

\author{Jiahui Lei}
\affiliation{%
  \institution{University of California, Berkeley}
  \country{USA}
}
\email{leijh@berkeley.edu}

\author{Zhiyang Dou}
\affiliation{%
  \institution{Massachusetts Institute of Technology}
  \country{USA}
}
\email{frankdou@mit.edu}

\author{Chenyue Cai}
\affiliation{%
  \institution{Princeton University}
  \country{USA}
}
\email{cc4880@princeton.edu}

\author{Chaoyue Song}
\affiliation{%
  \institution{Nanyang Technological University}
  \country{Singapore}
}
\email{chaoyue002@e.ntu.edu.sg}

\author{Adam Finkelstein}
\affiliation{%
  \institution{Princeton University}
  \country{USA}
}
\email{af@princeton.edu}

\author{Szymon Rusinkiewicz}
\affiliation{%
  \institution{Princeton University}
  \country{USA}
}
\email{smr@princeton.edu}

\renewcommand{\shortauthors}{Mou et al.}

\begin{abstract}
    Recent advances in automatic rigging now deliver animation-ready 3D assets at scale, yet generating the motion to drive them remains a bottleneck. Existing learned animators are topology-constrained: they rely on category-specific templates or require per-skeleton fine-tuning and reference motions at inference.
    We present \themodel{}, a unified foundation model that synthesizes articulated motion for arbitrary skeletons from a rigged 3D asset and a text prompt, with no test-time optimization or per-skeleton retraining.
    \themodel{} introduces a topology-aware diffusion transformer, which integrates skeletal topology into attention via three mechanisms: (1) a graph-aware attention bias from pairwise joint relations and geodesic distances; (2) a spectral rotary position embedding generalizing RoPE to arbitrary kinematic trees via the graph Laplacian; and (3) a global topological conditioner attention-pooled from the rest-pose skeleton.
    We also curate \thedata{}, 13,006 motion sequences spanning bipedal, quadrupedal, avian, marine, insectoid, serpentine, and articulated rigid objects with unified canonicalization and text pairing.
    Trained on this dataset, \themodel{} outperforms state-of-the-art baselines in quality, generalization, and efficiency, and supports zero-shot cross-topology transfer, in-betweening, expansion, and text-guided editing. \codeavailability
\end{abstract}

\ccsdesc[500]{Computing methodologies~Animation}
\ccsdesc[500]{Computing methodologies~Artificial intelligence}
\ccsdesc[500]{Computing methodologies~Machine learning}
\keywords{Motion Synthesis, Skeletal Animation, 3D Character Animation, Diffusion Models, Topology-Aware Learning}

\begin{teaserfigure}
  \includegraphics[width=\textwidth]{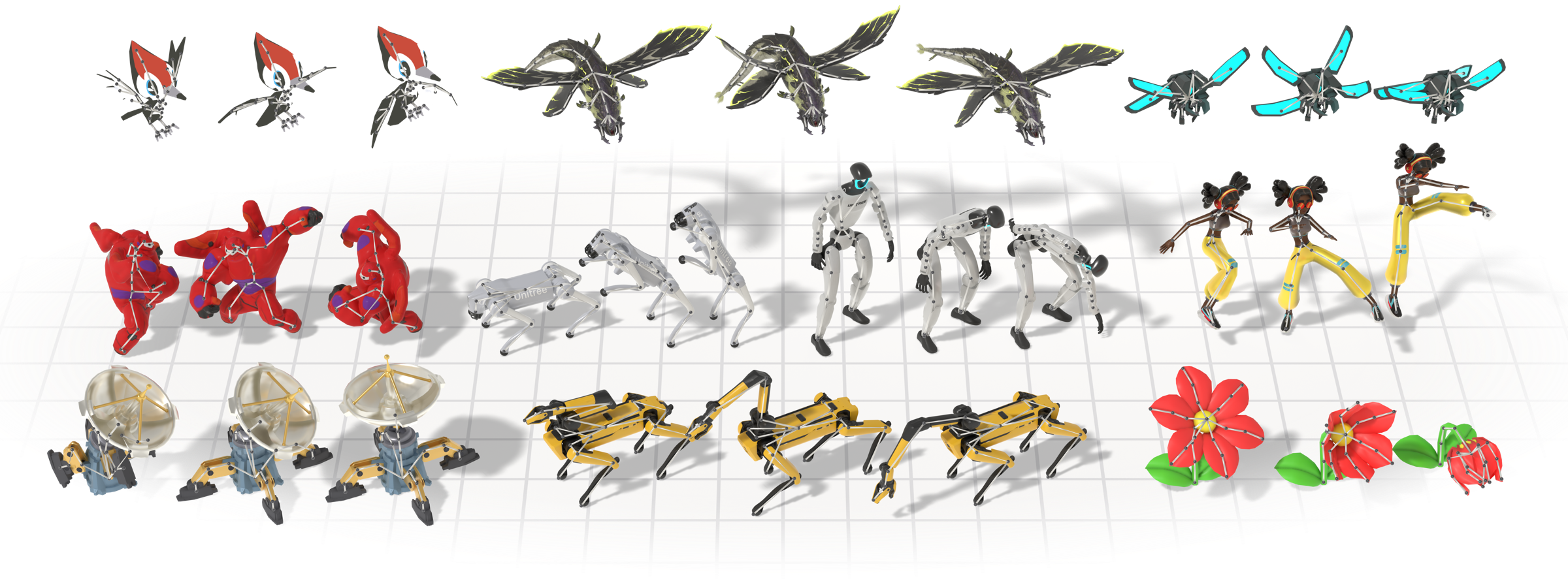}
  \Description{Examples of diverse rigged 3D characters and skeletons animated by UniMate from text prompts.}
  \caption{Given a rigged 3D asset and a text prompt, \themodel{} generates animations for characters with \emph{diverse skeletal topologies} within a \emph{single unified} model.}
  \label{fig:teaser}
\end{teaserfigure}

\maketitle

\section{Introduction}
\label{sec:introduction}

Character animation is fundamental to 3D content creation, central to film, gaming, virtual reality, and robotics simulation. Traditional pipelines define a \emph{unique} skeleton per character, over which artists craft motion through manual keyframing, motion capture, or slow per-asset optimization. Recent advances in 3D content creation~\cite{instant3d,zero123,one2345,pointe,trellis,hunyuan3d,rodin} and automatic rigging~\cite{puppeteer,riganything,unirig,rignet,magicarticulate} now deliver skeleton-ready 3D assets across a broad range of categories, from humans and animals to articulated rigid objects. Yet while \emph{rigged assets} can be generated at scale, the motion that drives them cannot---animation remains the laborious final bottleneck in an otherwise automated 3D content creation pipeline.

The bottleneck lies in the design of existing learned animators. State-of-the-art motion generators~\cite{mdm,modi,gmd,closd,hymotion,dart,actor,kimodo} are typically topology-constrained, relying on fixed category-specific skeleton templates such as SMPL~\cite{smpl} or SMAL~\cite{smal}. Topology-agnostic models~\cite{ganimator,sinmdm,anytop} relax these constraints but still require per-skeleton fine-tuning or reference motions at inference time. Mesh-based methods avoid explicit skeleton modeling altogether, but either depend on costly per-asset distillation~\cite{animate3d,motiondreamer,v2m4} or regress kinematically unconstrained vertex-wise deformations~\cite{gvfd,dnf,animateanymesh,driveanymesh}. These limitations motivate a unified foundation model that can synthesize motion for arbitrary skeletal topologies directly from high-level descriptions. Such a model would animate any rig in a single feed-forward pass and share motion priors across topologies, generalizing to unseen rigs and supporting a range of downstream applications (\cref{fig:motion-transfer,fig:motion-in-betweening,fig:motion-editing,fig:motion-expansion}).

However, developing such a unified animator poses two fundamental challenges. 
The first is \emph{modeling}: real-world skeletons are highly heterogeneous---bipedal humans, multi-legged insects, winged animals, and articulated rigid objects all exhibit distinct kinematic trees, joint counts, and motion patterns. This challenge is further compounded by the diverse motion behaviors associated with different morphologies. A general-purpose model must therefore treat skeletal topology as an explicit input, rather than baking it into an architectural prior, and reason jointly over structure and motion. 
The second is \emph{data}: text-paired motion corpora spanning diverse skeletal topologies remain scarce, with existing benchmarks dominated by humans~\cite{humanml3d,amass,kit} and a limited number of quadrupeds~\cite{omnimotiongpt}. Meanwhile, raw rigged 4D assets~\cite{truebones,objaverse,objaversexl} are often noisy and inconsistent, and lack unified preprocessing and canonicalization across topologies, leaving data-driven approaches without coherent supervision for cross-topology generalization.

In this work, we propose \emph{\textbf{\themodel{}}}, a unified foundation model that animates diverse skeletons. Given a rigged 3D asset and a natural-language prompt, \themodel{} synthesizes plausible articulated motion with no test-time fitting or per-skeleton specialization (see \cref{fig:teaser}). Joint training across a wide range of skeletons lets the model learn motion patterns that are shared and transferable across topologies, enabling stronger generalization to unseen rigs and motion transfer between heterogeneous structures.

At the core of \themodel{} is the Topology-Aware Diffusion Transformer (TADiT), a flow-matching architecture in which attention layers jointly reason over rest-pose kinematics and motion manifolds through a shared token stream. To encode heterogeneous topologies, we equip TADiT with three key design choices.
First, vanilla self-attention is blind to the underlying kinematic graph. We therefore inject a \emph{graph-aware attention bias}~\cite{graphormer} derived from pairwise joint relations and geodesic distances, so anatomically nearby joints attend more strongly while the model retains its capacity for long-range, full-body coordination.
Second, we introduce \emph{Spec-RoPE}, a spectral rotary position embedding that generalizes RoPE~\cite{rope} to arbitrary kinematic trees by deriving rotary angles from the graph Laplacian spectrum. With provable translation invariance in spectral coordinates and equivariance under joint permutation, Spec-RoPE adapts to skeletons of varying size and connectivity---a property that index- or coordinate-based encodings cannot provide.
Third, a \emph{global topological conditioner}, attention-pooled from the skeleton tokens, modulates every transformer block through AdaLN-Zero~\cite{dit}, so layer-wise feature statistics adapt to the input skeleton and provide global structural context complementing the local signals above.

To support training at scale, we curate \thedata{}, a heterogeneous motion dataset of 13{,}006 animation sequences (roughly 20 hours) drawn from Truebones~\cite{truebones}, Mixamo~\cite{mixamo}, and Objaverse-XL~\cite{objaverse,objaversexl}, pairing thousands of distinct rigs across bipedal, quadrupedal, avian, marine, insectoid, serpentine, and articulated rigid objects with 3{,}584 unique text prompts that span a broad action spectrum, including locomotion, combat, idle, mechanical articulation, and object manipulation. Rigorous filtering followed by unified canonicalization yields a shared representation across skeleton types, while online skeletal augmentation further broadens topological coverage during training. This scale and coverage substantially exceed those of prior work~\cite{anytop} and are essential for cross-skeleton generalization.

Extensive experiments demonstrate that \themodel{} achieves state-of-the-art performance on topology-agnostic motion generation and mesh animation, surpassing prior methods in quality, generalization, and efficiency. \themodel{} also supports zero-shot downstream tasks such as cross-topology motion transfer, in-betweening, expansion, and text-guided editing, serving as a controllable engine for scalable 3D character animation. In summary, our contributions are:

\begin{itemize}[leftmargin=1em, labelsep=0.5em, itemsep=0.2em, topsep=0.2em]
  \item We present \themodel{}, a unified foundation model that synthesizes articulated motion for skeletons of arbitrary topology from a rigged 3D asset and a text prompt.
  \item We propose TADiT, which couples motion and skeletal structure within shared attention layers through a graph-aware attention bias, the Spec-RoPE spectral rotary position embedding with provable structural properties, and a global topological conditioner.
  \item To facilitate training and benchmarking, we curate \thedata{}, comprising 13{,}006 motion sequences over thousands of skeletons spanning bipedal, quadrupedal, avian, marine, insectoid, serpentine, and articulated rigid objects, unified by a canonicalization pipeline and online skeletal augmentation.
  \item We conduct various experiments showing that \themodel{} improves over prior methods in quality, generalization, and runtime, and enables zero-shot cross-topology motion transfer, in-betweening, expansion, and real-time text-guided editing.
\end{itemize}

\begin{figure*}[t]
  \centering
      \includegraphics[width=1.0\linewidth]{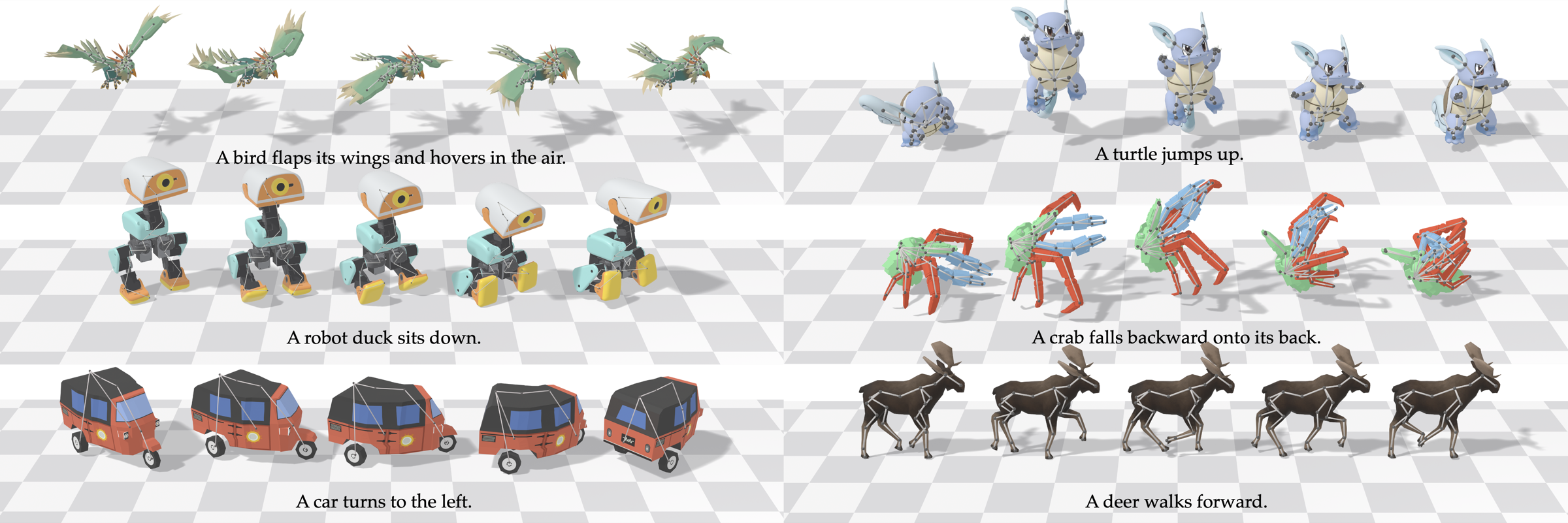}
  \Description{Qualitative UniMate animation sequences for characters with heterogeneous skeletons and motion prompts.}
  \caption{\textbf{Animations generated by \themodel{}.} Our method generalizes across heterogeneous skeletons and diverse motion prompts.}
  \label{fig:qualitative-results}
\end{figure*}

\section{Related Work}
\label{sec:related-work}

\subsection{3D Animation}
\label{sec:related-3d-animation}

A growing body of work animates 3D assets by distilling image- and video-generative priors or by reconstructing motion from generated videos~\cite{diffusion4d,animate3d,motiondreamer,animax,anymole,dimo,chord,v2m4}.
These methods typically require costly per-asset test-time optimization and tend to produce jittery, unstable trajectories, making them ill-suited to large-scale or interactive use.
A second class of methods~\cite{gvfd,dnf,animateanymesh,shapegen4d,bimotion,driveanymesh,motion3to4,actionmesh} trains feed-forward networks that regress per-vertex or per-point deformations directly.
Operating in raw geometry space sacrifices the compactness of skeletal representations and breaks native compatibility with the rig-driven ecosystem: linear blend skinning~\cite{lbs,neural-blend-shapes}, physics-based controllers~\cite{closd}, physics simulators~\cite{mujoco,isaac}, and motion-capture pipelines~\cite{mocapanything}.
Recent advances in automatic rigging~\cite{puppeteer,riganything,unirig,rignet,magicarticulate} have made high-quality skeletons widely accessible, yet driving the resulting rigs still demands manual keyframing or expensive per-asset optimization~\cite{akd,puppeteer,animamimic}.
\themodel{} closes this gap with a data-driven generative model that operates directly on arbitrary skeletons.

\subsection{Cross-Topology Motion Generation and Retargeting}
\label{sec:related-cross-topology}

The dominant family of learned motion generators, from human motion models~\cite{mdm,modi,aamdm,acmdm,gmd,lora-mdm,camdm,momo,closd,priormdm,hymotion,dart,case,emdm,kimodo,tlcontrol,gotozero,scamo} to species-specific animal models~\cite{ponymation,animo,x-mogen}, assumes a single fixed skeleton template, typically inherited from parametric body models such as SMPL~\cite{smpl,smpl-x} and SMAL~\cite{smal}, and therefore cannot handle characters whose topology departs from the template.

A separate family of example-based methods sidesteps neural-network training by stitching patches from exemplar motion clips: generative motion matching~\cite{example} carries the patch nearest-neighbor synthesis of Drop-the-GAN~\cite{dropgan} over to motion, and Motion2Motion~\cite{motion2motion} extends it to cross-topology transfer through sparse correspondences. Although topology-flexible, these methods still require an exemplar motion for every target and cannot synthesize motion from a static rigged asset alone.
Closer to our setting, GANimator~\cite{ganimator} and SinMDM~\cite{sinmdm} learn neural generators on arbitrary topologies, but train a separate model per skeleton and so do not generalize across structures.
Cross-topology retargeting~\cite{pool,pose2motion,palum,walkthedog,same} bypasses the template constraint from a different angle, but again only by transferring an existing source motion onto a target skeleton.
The closest prior work, AnyTop~\cite{anytop}, jointly trains a single diffusion model over heterogeneous animal skeletons, but is limited to a small animal corpus~\cite{truebones}, requires motion data of the target skeleton at inference to estimate its normalization statistics, and offers no text conditioning.

In contrast, a single \themodel{} model covers a far broader range of skeletons, from humans and animals to general articulated objects, accepts text conditioning, and animates a rigged mesh end-to-end, without reference motion or per-skeleton training.

\begin{figure*}[t]
  \centering
  \includegraphics[width=1.0\linewidth]{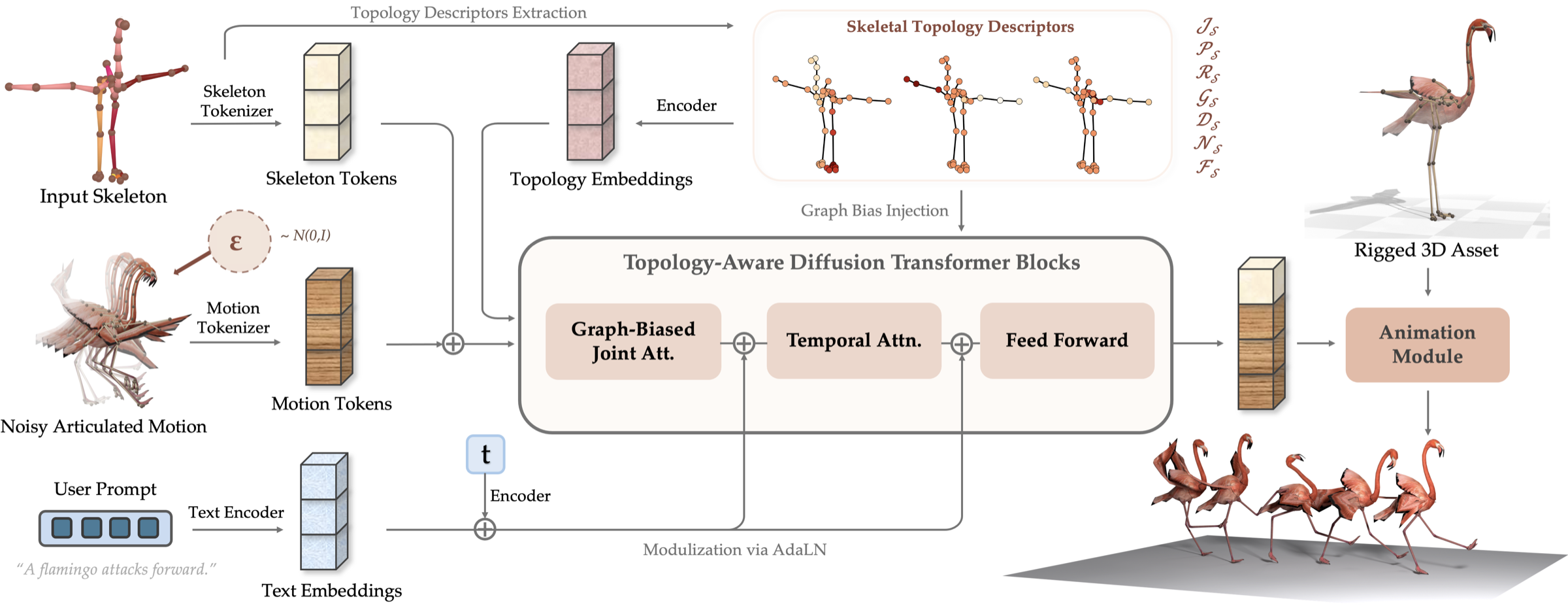}
  \Description{UniMate pipeline from a rigged mesh and text prompt through topology-aware joint tokens to generated skeletal animation.}
  \caption{\textbf{\themodel{} pipeline.}
  The proposed TADiT operates on a joint token space that combines per-frame motion features with rest-pose skeletal descriptors, while injecting skeletal graph-structured bias into both positional encoding and attention computation. Conditioned on input text prompts, our unified model produces realistic, coherent animations while demonstrating strong cross-topology generalization and efficient runtime performance.}
  \label{fig:method-overview}
\end{figure*}

\section{Method}
\label{sec:method}

Given a rigged 3D asset and a text prompt, our goal is to synthesize a plausible motion sequence that animates the input mesh (\cref{fig:method-overview}). We first introduce a unified representation for heterogeneous skeletons and motion sequences (\cref{sec:representation}), then present our Topology-Aware Diffusion Transformer (\cref{sec:architecture}), and finally describe the training objective and inference procedure (\cref{sec:training}).

\subsection{Skeleton and Motion Representation}
\label{sec:representation}
An articulated rigged 3D asset is animated by a \emph{skeleton}---a kinematic tree with a joint hierarchy and bone lengths, whose joints drive mesh deformation through forward kinematics. A \emph{rest pose} of a skeleton is the neutral undeformed reference configuration. Our model takes as input the skeleton at rest pose and builds a diffusion model over a motion sequence defined on it.

\boldstartspace{Skeleton Definition.}
We model an articulated object as a rooted kinematic tree with $J$ joints. To represent skeletons with heterogeneous topologies in a shared transformer space, we first assign each skeleton a canonical joint ordering. Concretely, we linearize the tree using breadth-first search (BFS) given the known root node
\begin{equation}
\mathcal{J}_{\mathcal{S}}
=
\bigl[(j_1,\mathrm{pa}_1),\,(j_2,\mathrm{pa}_2),\,\dots,\,(j_J,\mathrm{pa}_J)\bigr],
\end{equation}
where $j_k\in\mathbb{R}^3$ denotes the rest-pose position of the $k$-th joint and $\mathrm{pa}_k\in\{1,\dots,J\}$ denotes its parent index. The BFS ordering places the root at $k=1$; since it has no parent, we adopt the self-parent convention $\mathrm{pa}_1=1$.

\boldstartspace{Topology-Diameter Normalization.}
Skeletons in our dataset vary significantly in absolute size and topological extent. To place them into a comparable representation space, we normalize each rest-pose skeleton by its topology diameter, defined as
\begin{equation}
d_{\mathrm{topo}}
=
\max_{i,j}\,\mathrm{dist}^{\mathrm{geo}}_{\mathrm{tree}}(i,j),
\end{equation}
where $\mathrm{dist}^{\mathrm{geo}}_{\mathrm{tree}}(i,j)$ denotes the geodesic distance between joints $i$ and $j$ along the kinematic tree. This normalization removes scale differences while preserving the relative kinematic structure.

  \begin{figure*}[t]
  \centering
      \includegraphics[width=1.0\linewidth]{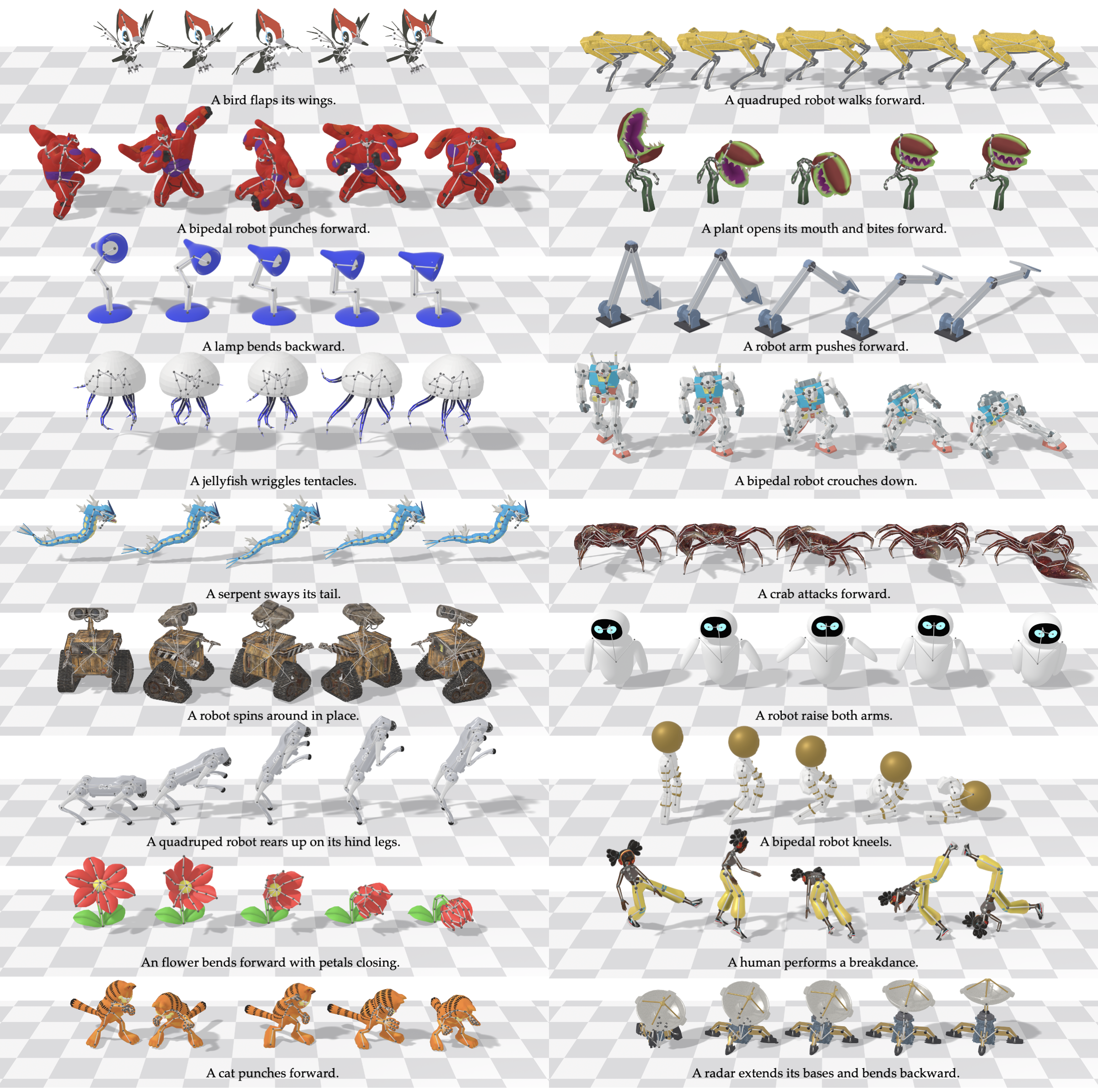}
  \Description{Animation sequences generated for quadrupedal, bipedal, plant-like, avian, and articulated-object skeletons from text prompts.}
  \caption{\textbf{Cross-topology animation.} \themodel{} generates prompt-aligned motions for diverse characters and articulated objects within a single unified model.}
  \label{fig:cross-topology-animations}
\end{figure*}

\boldstartspace{Local Topology Descriptors.}
Given the canonicalized skeleton, we describe its rest-pose geometry and local topology using several descriptors. Rest-pose joint positions are stacked in $\mathcal{P}_{\mathcal{S}}\in\mathbb{R}^{J\times 3}$. Local kinematic context is encoded by a pairwise relation matrix $\mathcal{R}_{\mathcal{S}}\in\mathbb{N}_0^{J\times J}$, whose entries specify relation types such as parent, child, sibling, or ancestor; a pairwise graph-distance matrix $\mathcal{G}_{\mathcal{S}}\in\mathbb{N}_0^{J\times J}$ on the kinematic tree; a per-joint depth vector $\mathcal{D}_{\mathcal{S}}\in\mathbb{N}_0^{J}$; and a per-joint name index $\mathcal{N}_{\mathcal{S}}\in\mathbb{N}_0^{J}$ into a joint-name vocabulary that captures semantic identity.

\begin{figure}[b]
  \centering
  \includegraphics[width= 1.0\linewidth]{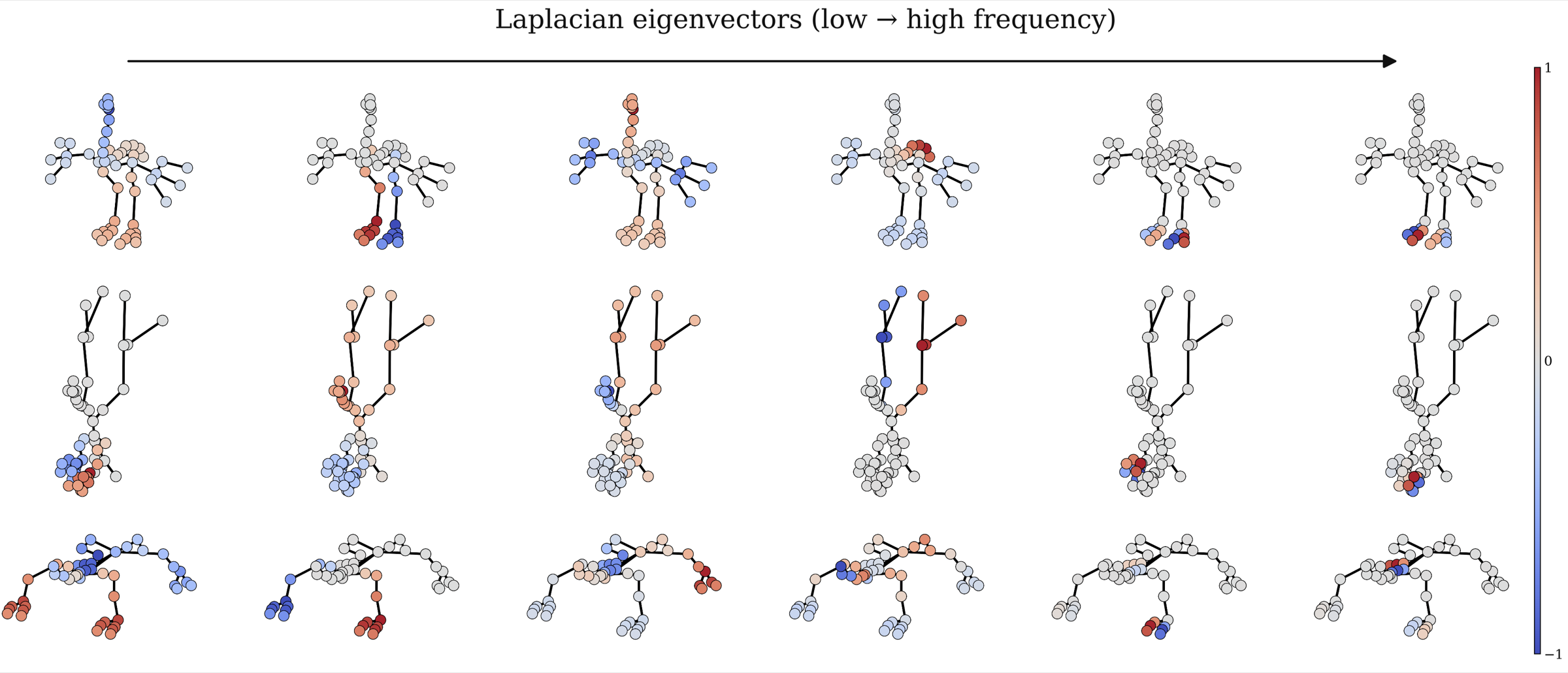}
  \Description{Skeleton joints colored by graph Laplacian eigenvectors from low to high spatial frequency.}
  \caption{\textbf{Spectral visualization.} From left to right, Laplacian eigenvectors increase in frequency. Low-frequency modes capture global kinematic structure, while higher-frequency modes encode finer local relationships.}
  \label{fig:spectral-visualization}
\end{figure}

\boldstartspace{Spectral Coordinates.}
To complement the discrete topological descriptors above with a continuous encoding of skeletal structure, we additionally compute spectral features from the kinematic graph. Let $A\in\{0,1\}^{J\times J}$ denote the adjacency matrix of the skeleton and $L_{\mathcal{S}}=\mathrm{diag}(A\mathbf{1})-A$ its graph Laplacian. Let $L_{\mathcal{S}} = U \Lambda U^\top$ be its eigendecomposition. Discarding the trivial constant eigenvector $u_0$, we define the spectral feature of joint $j$ using the first $m$ non-trivial eigenvectors:
\begin{equation}
\mathcal{F}_{\mathcal{S}}(j)
=
[u_1(j),\,u_2(j),\,\dots,\,u_m(j)] \in \mathbb{R}^m.
\end{equation}
These spectral coordinates provide a continuous encoding of joint location on the kinematic graph (see \cref{fig:spectral-visualization}), complementing the discrete relation and distance descriptors.

Overall, we represent a skeleton as
\begin{equation}
\mathcal{S}
=
\{
\mathcal{J}_{\mathcal{S}},
\mathcal{P}_{\mathcal{S}},
\mathcal{R}_{\mathcal{S}},
\mathcal{G}_{\mathcal{S}},
\mathcal{D}_{\mathcal{S}},
\mathcal{N}_{\mathcal{S}},
\mathcal{F}_{\mathcal{S}}
\},
\end{equation}
where $\mathcal{J}_{\mathcal{S}}$ specifies the ordered kinematic tree, $\mathcal{P}_{\mathcal{S}}$ the rest-pose geometry, $\mathcal{R}_{\mathcal{S}}, \mathcal{G}_{\mathcal{S}}, \mathcal{D}_{\mathcal{S}}$ the discrete topological structure, $\mathcal{N}_{\mathcal{S}}$ the semantic joint identity, and $\mathcal{F}_{\mathcal{S}}$ the global spectral descriptor.

\boldstartspace{Motion Representation.}
Given a skeleton $\mathcal{S}$, a motion sequence is $X = \{\mathbf{m}_j^t\} \in\mathbb{R}^{T\times J\times D}$ over $T$ frames and $J$ joints. Following~\cite{humanml3d,anytop}, each joint feature has dimension $D=12$:
\begin{equation}
\mathbf{m}_j^t = [\mathbf{p}_j^t,\, \mathbf{r}_j^t,\, \mathbf{v}_j^t] \in \mathbb{R}^{D}.
\end{equation}

\boldstartspace{$\mathbf{p}_j^t\in\mathbb{R}^3$: Position.}
For non-root joints, horizontal $(x,z)$ components are taken relative to the root and rotated into the facing-canonical frame, with vertical $y$ kept in the world frame. The root joint $\mathbf{p}_1^t$ retains its global position.

\boldstartspace{$\mathbf{r}_j^t\in\mathbb{R}^6$: Rotation.}
Represented in the continuous 6D format~\cite{6drot}.

\boldstartspace{$\mathbf{v}_j^t\in\mathbb{R}^3$: Velocity.}
Computed in the world frame as the temporal derivative of global joint positions, preserving absolute motion cues that the canonical-frame projection of $\mathbf{p}_j^t$ would otherwise discard.

  \begin{figure*}[t]
  \centering
  \setlength{\belowcaptionskip}{\floatsep}
  \includegraphics[width= 1.0\linewidth]{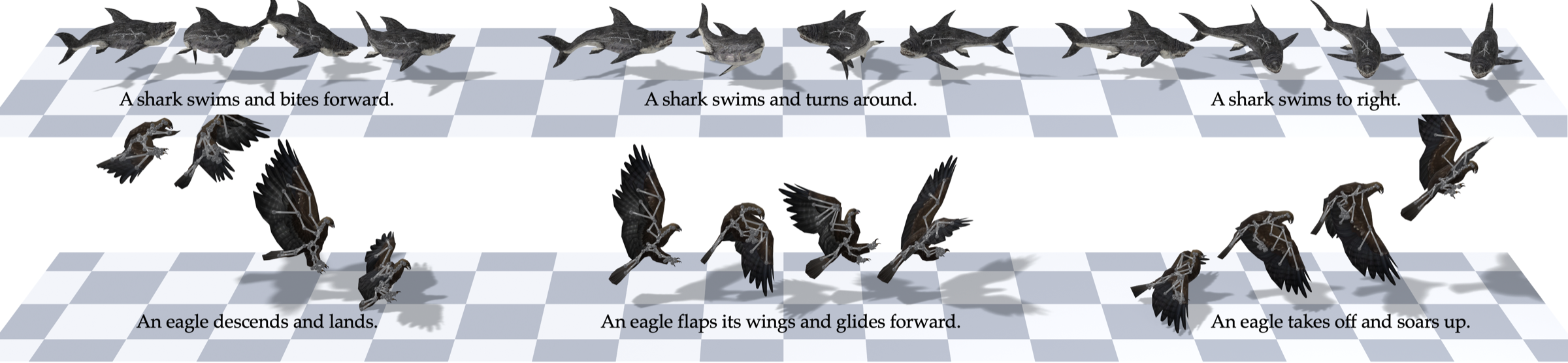}
  \Description{One character skeleton animated according to several different text prompts.}
      \caption{\textbf{One skeleton, diverse prompts.} Given a single skeleton, \themodel{} synthesizes distinct, prompt-faithful motions for different input text prompts.}
  \label{fig:diverse-prompts}

  \setlength{\belowcaptionskip}{0pt}
  \includegraphics[width= 1.0\linewidth]{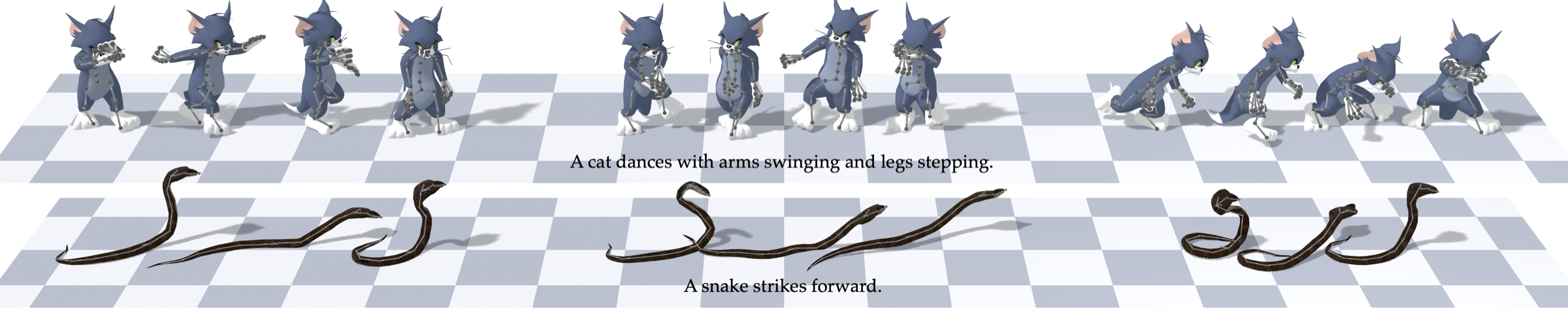}
  \Description{Several distinct motion samples generated for the same character skeleton and text prompt.}
  \caption{\textbf{One skeleton, one prompt, diverse motions.} Given the same skeleton and text prompt, \themodel{} generates diverse plausible motion samples.}
  \label{fig:diverse-motions}
\end{figure*}

  \begin{figure*}[t]
  \centering
  \includegraphics[width=1.0\linewidth]{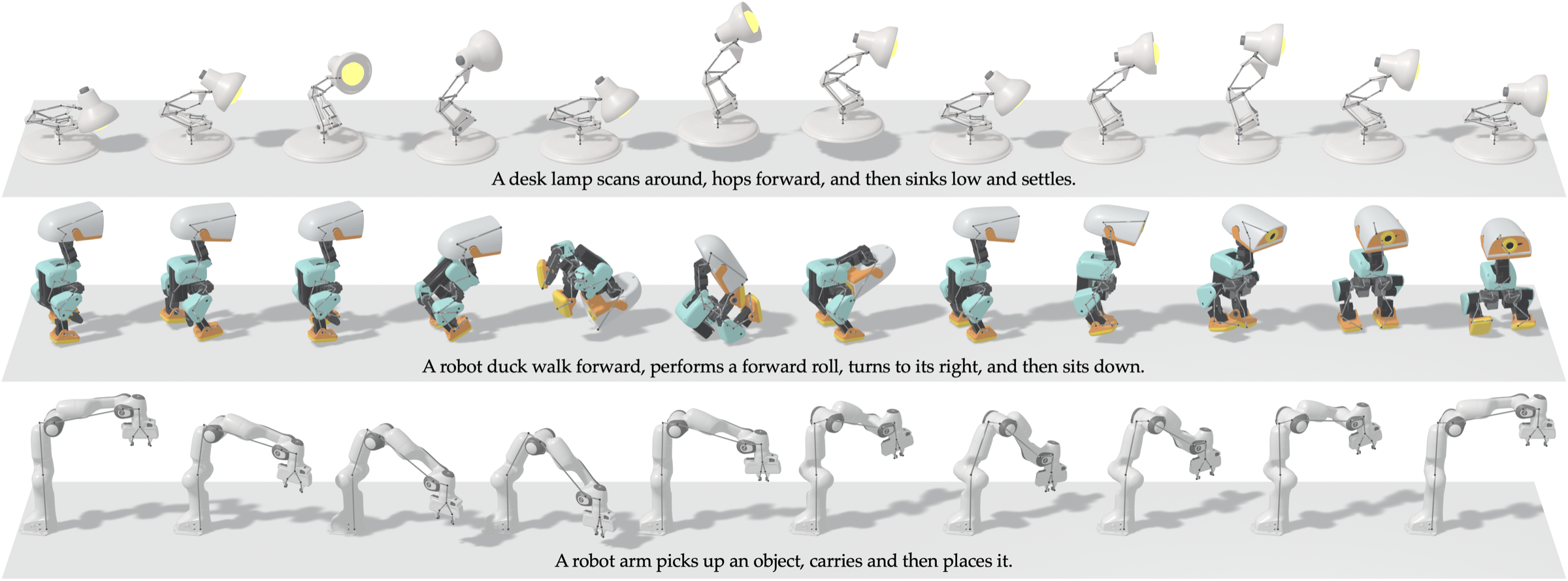}
  \Description{Three rows of eleven keyframes, each from a single generated 180-frame sequence: a desk lamp scanning around, hopping forward, and settling low; a robot duck walking forward, performing a forward roll, turning right, and sitting down; and a robot arm picking up an object, carrying it, and placing it.}
      \caption{\textbf{Long-horizon generation.} A variant trained at 180 frames generates long sequences that stay temporally coherent and drift-free.}
  \label{fig:motion-long}
\end{figure*}

\subsection{Topology-Aware Diffusion Transformer (TADiT)}
\label{sec:architecture}

Our goal is to generate a plausible motion sequence $\hat{x}_{1}$ conditioned on a rest-pose skeleton $\mathcal{S}$ and a text prompt $c$. This is challenging because the model must generalize across heterogeneous skeletons with different kinematic trees, joint counts, and motion patterns. To address this, we introduce the \emph{Topology-Aware Diffusion Transformer} (TADiT), which injects topology through a graph-aware attention bias, a spectral rotary position embedding (Spec-RoPE), and a global topological conditioner. We train TADiT with conditional flow matching~\cite{flow}, learning a velocity field $v_\theta(x_\tau,\tau,c,\mathcal{S})$ that transforms Gaussian noise $x_0\sim\mathcal{N}(0,I)$ into a valid motion sequence $x_1$.

\boldstartspace{Skeleton and Motion Tokenization.}
For each joint $j$, we form the skeleton token $\mathbf{T}_j \in \mathbb{R}^d$ by concatenating MLP-projected rest-pose positions of the joint and its parent and applying a fusion MLP:
\begin{equation}
\mathbf{T}_j = \mathrm{MLP}_{\mathrm{fuse}}\Bigl(\mathrm{Concat}\bigl(\mathrm{MLP}_{\mathrm{jt}}(\mathcal{P}_{\mathcal{S}}(j)),\, \mathrm{MLP}_{\mathrm{pa}}(\mathcal{P}_{\mathcal{S}}(\mathrm{pa}_j))\bigr)\Bigr),
\end{equation}
where $\mathcal{P}_{\mathcal{S}}(j)$ is the rest-pose position of joint $j$.

Each motion feature $\mathbf{m}_j^t \in \mathbb{R}^{D}$ (defined in \cref{sec:representation}) is projected to dimension $d$ and augmented with a learnable depth embedding and a joint-name text embedding as hierarchical and semantic priors:
\begin{equation}
\mathbf{M}_j^t = \mathrm{MLP}(\mathbf{m}_j^t) + E_{\mathrm{depth}} \bigl(\mathcal{D}_{\mathcal{S}}(j)\bigr) + E_{\mathrm{name}} \bigl(\mathcal{N}_{\mathcal{S}}(j)\bigr).
\end{equation}

We prepend the skeleton tokens to the motion tokens along the temporal axis, forming $Z=\mathrm{Concat}(\mathbf{T},\mathbf{M})\in\mathbb{R}^{(T+1)\times J\times d}$, such that every transformer block operates on a unified token space containing both static skeletal structure and dynamic per-frame motion.

\boldstartspace{Skeletal-Temporal Transformer Blocks.}
Each block operates on $Z\in\mathbb{R}^{(T+1)\times J\times d}$ and is conditioned on the diffusion timestep, text prompt, and global topology embeddings. For tractable cost on heterogeneous skeletons, each block uses a factorized attention with a \emph{joint} branch (across joints at each frame) and a \emph{temporal} branch (across frames at each joint), followed by a feed-forward sublayer:
\begin{equation}
Z^{(\ell+1)}
=
\mathrm{FFN}\Big(
\mathrm{TempAttn}\big(
\mathrm{JointAttn}(Z^{(\ell)}, \mathcal{S}),\, c
\big),\, c
\Big).
\end{equation}
The temporal branch is a standard multi-head self-attention with 1D rotary position embedding (RoPE)~\cite{rope} on the frame index. Kinematic structure is exposed exclusively to the joint branch, through a graph-aware attention bias and the Spectral Rotary Position Embedding (Spec-RoPE) described next.

\boldstartspace{Graph-Aware Attention Bias.}
We inject the kinematic graph into attention through a learned bias added to the joint-attention logits, exposing pairwise structural relations that are hard for vanilla self-attention to recover from token features alone. Following~\citet{graphormer}, the pairwise graph-distance and relation descriptors $\mathcal{G}_{\mathcal{S}},\mathcal{R}_{\mathcal{S}}$ are embedded by lookup tables
\begin{equation}
e_d(i,j)=E_d\!\left(\mathcal{G}_{\mathcal{S}}(i,j)\right), \qquad
e_r(i,j)=E_r\!\left(\mathcal{R}_{\mathcal{S}}(i,j)\right),
\end{equation}
and projected to a per-head scalar bias
\begin{equation}
B_{ij}^{(h)} = (w_d^{(h)})^{\!\top} e_d(i,j) + (w_r^{(h)})^{\!\top} e_r(i,j),
\end{equation}
with head-specific projection vectors $w_d^{(h)}, w_r^{(h)} \in \mathbb{R}^{d_e}$. Letting $\tilde{q}_i^{(h)},\tilde{k}_j^{(h)} \in \mathbb{R}^{d_h}$ denote the Spec-RoPE-rotated queries and keys for joints $i$ and $j$, the joint-attention logits read
\begin{equation}
\mathrm{Attn}^{(h)}(i,j)
=
{\textstyle\frac{\textstyle 1}{\sqrt{d_h\!}}} \bigl(\tilde{q}_i^{(h)}\bigr)^{\!\top}\, \tilde{k}_j^{(h)}
+ B_{ij}^{(h)}.
\end{equation}
The bias is shared across frames, so its memory cost is independent of sequence length. Because it is parameterized by graph-distance and relation-type embeddings rather than absolute joint indices, it transfers to unseen topologies (\cref{sec:ablation}).

\boldstartspace{Spectral Rotary Position Embedding (Spec-RoPE).}
Kinematic trees have no canonical ordering, so the index used by 1D RoPE is ill-defined for joints. We instead derive rotary angles from the spectrum of the graph Laplacian~\cite{graph-transformer,graphgps}, applied to the joint branch only:
\begin{equation}
\underbrace{\boldsymbol{\theta}_j = \boldsymbol{\omega}\, j}_{\text{1D RoPE}}
\;\;\longrightarrow\;\;
\underbrace{\boldsymbol{\theta}_j = f(\mathbf{s}_j)}_{\text{Spec-RoPE}},
\end{equation}
where $j\in\mathbb{N}$ is the token index, $\boldsymbol{\omega}\in\mathbb{R}^{d_h/2}$ the standard frequency vector, $\mathbf{s}_j = \mathcal{F}_{\mathcal{S}}(j) \in\mathbb{R}^{m}$ the joint's spectral coordinate from \cref{sec:representation}, and $f$ a learned angle map specified below.

\noindent
\emph{Intuition.} RoPE relies on positional coordinates to define relative phase offsets in attention. For temporal tokens, the frame index is a natural causal position; for kinematic-tree joints, the BFS index is arbitrary: two joints adjacent in the index can lie on opposite limbs. The spectral coordinate $\mathbf{s}_j$ replaces it with an intrinsic position on the graph: the low-frequency Laplacian eigenvectors capture the coarse global organization of the kinematic tree, while higher-frequency eigenvectors progressively encode finer structural variation (\cref{fig:spectral-visualization}). The rotary phase therefore depends on \emph{where a joint sits on the skeleton}, not on how it is serialized.

The spectral coordinate $\mathbf{s}_j = \mathcal{F}_{\mathcal{S}}(j) \in \mathbb{R}^{m}$ comprises the leading $m$ non-trivial eigenvectors of $L_{\mathcal{S}}$. Since these are determined only up to sign, we realize $f$ as a SignNet~\cite{signnet}:
\begin{equation}
\boldsymbol{\theta}_j
=
\mathrm{MLP}_{\mathrm{ang}}\!\left(
\mathrm{Concat}\!\left(
\bigl\{\mathrm{MLP}_{\mathrm{sym}}(u_n(j)) + \mathrm{MLP}_{\mathrm{sym}}(-u_n(j))\bigr\}_{n=1}^{m}
\right)
\right)
\!.
\end{equation}
The angles $\boldsymbol{\theta}_j \in \mathbb{R}^{d_h/2}$ then drive the standard RoPE block-diagonal rotation $\mathbf{R}(\boldsymbol{\theta}_j)$~\cite{rope}, applied to per-joint queries and keys as $\tilde{q}_j = \mathbf{R}(\boldsymbol{\theta}_j)\, q_j$ and $\tilde{k}_j = \mathbf{R}(\boldsymbol{\theta}_j)\, k_j$.

Spec-RoPE satisfies two structural properties, formalized in \appendixref{app:spectral-rope-theory}: \emph{translation invariance in spectral coordinates} and \emph{equivariance under joint permutation}. Together, these allow Spec-RoPE to adapt to skeletons of varying size and connectivity. We empirically validate the effect of Spec-RoPE in \cref{sec:ablation}.

\boldstartspace{Global Topological Conditioner.}
Beyond the pairwise graph-aware attention bias and per-joint Spec-RoPE topology signals, each transformer block also requires a single, joint-count-invariant skeleton summary. We obtain this global topological condition $c_{\mathrm{topo}}$ via \emph{attention pooling}~\cite{set-transformer} over the skeleton token sequence $\mathbf{T}$, producing a fixed-size summary independent of joint count and, after the final mean aggregation, of joint ordering. A small set of $n_q$ learnable query tokens $\mathbf{Q}_{\mathrm{pool}} \in \mathbb{R}^{n_q \times d}$ serves as a content-adaptive readout: each query attends to all skeleton tokens through cross-attention, with $\mathbf{T}$ providing the keys and values,
\begin{equation}
\mathbf{H} = \mathrm{softmax}\left({\textstyle\frac{\textstyle 1}{\sqrt{d}}}(\mathbf{Q}_{\mathrm{pool}} W_q)(\mathbf{T} W_k)^\top
\right) \mathbf{T}\, W_v,
\end{equation}
where $W_q, W_k, W_v \in \mathbb{R}^{d\times d}$ are learnable projections. The query outputs $\mathbf{H} \in \mathbb{R}^{n_q\times d}$ are aggregated by mean pooling into $c_{\mathrm{topo}}$, which is fused with the timestep and text embeddings and injected into every block via AdaLN-Zero~\cite{dit} (detailed in \appendixref{app:conditioning}), yielding a holistic skeletal context.

\subsection{Training and Inference}
\label{sec:training}

\boldstartspace{Training Objective.}
We supervise $v_\theta$ with three complementary losses, with full expressions deferred to \appendixref{app:training-objective}.

\emph{Flow-matching MSE.} The base loss $\mathcal{L}_{\mathrm{mse}}$ is a masked mean-squared error against the target velocity $v^{*}=x_1-x_0$, with padded joint slots zeroed out in heterogeneous-skeleton batches.

\emph{Geodesic rotation loss.} Because joint rotations live on the non-Euclidean manifold $SO(3)$, an isotropic MSE on their 6D channels is geometrically misaligned. We therefore add a geodesic loss $\mathcal{L}_{\mathrm{geo}}$ on the one-step denoised rotations $\hat{R}_j^t \in SO(3)$ from $\hat{x}_1=x_\tau+(1-\tau)\,v_\theta$, which penalizes their angular deviation from the ground truth.

\emph{Velocity smoothness.} To suppress high-frequency jitter, a smoothness regularizer $\mathcal{L}_{\mathrm{smooth}}$ penalizes the temporal acceleration of the denoised velocity channels of $\hat{x}_1$.

The final objective is a weighted combination of these three losses:
\begin{equation}
\mathcal{L}
=
\mathcal{L}_{\mathrm{mse}}
+ \lambda_{\mathrm{geo}}\,\mathcal{L}_{\mathrm{geo}}
+ \lambda_{\mathrm{smooth}}\,\mathcal{L}_{\mathrm{smooth}}.
\end{equation}

\boldstartspace{Inference.}
At inference time, we draw $x_0 \sim \mathcal{N}(0, I)$ and integrate the learned velocity field $\dot{x}_\tau = v_\theta(x_\tau, \tau, c, \mathcal{S})$ from $\tau=0$ to $1$ using a fixed-step Euler solver, yielding the generated motion features $\hat{x}_{1}$. Each step applies classifier-free guidance~\cite{cfg}.

\begin{figure*}[t]
  \centering
  \includegraphics[width=1.0\linewidth]{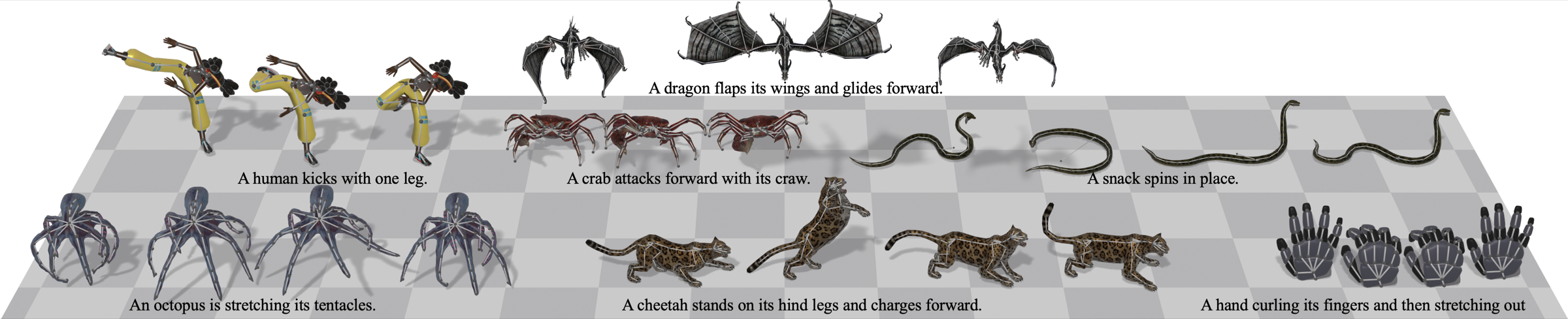}
  \Description{Representative UniML3D assets, skeletons, and motions across multiple morphology categories.}
  \caption{\textbf{Samples from the \thedata{} dataset.}
  Our dataset spans diverse skeletons across bipedal, quadrupedal, avian, marine, insectoid, serpentine, and articulated rigid objects, with detailed skeleton annotations and coherent text prompts paired with motion sequences.}
  \label{fig:dataset-overview}
\end{figure*}

\begin{figure}[b]
  \centering
  \includegraphics[width=\linewidth]{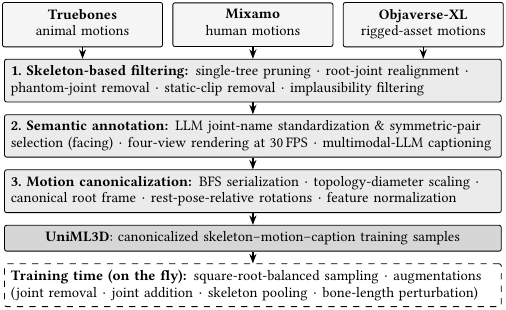}
  \Description{Flowchart of the UniML3D curation pipeline: three motion sources feed skeleton-based filtering, semantic annotation, and motion canonicalization, producing the dataset, which is consumed at training time with balanced sampling and augmentation.}
  \caption{\textbf{\thedata{} data-processing pipeline.} Source assets pass through skeleton-based filtering, semantic annotation, and motion canonicalization; balanced sampling and augmentation happen on the fly during training.}
  \label{fig:data-pipeline}
\end{figure}

\section{\thedata{} Dataset}
\label{sec:dataset}
Training a truly generalizable animation model for diverse object categories requires a large-scale dataset with varied skeletal structures and plausible motion sequences. However, raw 4D motion sources are noisy and inconsistent, often containing disconnected or scene-level skeletons, broken roots, non-functional joints, physically implausible motions, and mismatched coordinate frames or facing directions, making them unsuitable for direct cross-topology motion learning without rigorous preprocessing.

To address this, we curate \thedata{} from three complementary sources: Truebones~\cite{truebones}, Mixamo~\cite{mixamo}, and Objaverse-XL~\cite{objaverse,objaversexl}. After filtering and canonicalization, \thedata{} comprises 13{,}006 motion sequences and 2{,}140{,}232 frames over thousands of skeletons, spanning bipedal, quadrupedal, avian, marine, insectoid, serpentine, and articulated rigid objects (\cref{fig:dataset-overview}). \cref{fig:data-pipeline} summarizes this pipeline; per-source statistics and per-stage details are in \appendixref{app:dataset}.

\boldstartspace{Skeleton-Based Filtering.}
We apply a multi-stage filter so that every retained skeleton is connected and kinematically valid, and every retained clip carries plausible, non-trivial motion:
\begin{enumerate}
  \item \emph{Single-tree pruning.} Multiple or disconnected kinematic trees are pruned, keeping only the primary skeleton: the tree with the largest cumulative skinning weight.
  \item \emph{Root-joint realignment.} Spurious or misaligned root joints are corrected by propagating their global transformation onto the semantic root via forward kinematics.
  \item \emph{Phantom-joint removal.} Non-functional joints, e.g., IK controllers and helper bones with zero skinning weight, are recursively pruned to streamline the topology.
  \item \emph{Static-clip removal.} Clips with negligible activity, quantified by bone-length-normalized global displacement, are discarded.
  \item \emph{Implausibility filtering.} Clips with out-of-distribution root velocities or per-joint angular jitter above an anatomical threshold are discarded.
\end{enumerate}

\boldstartspace{Motion Canonicalization.}
Building on the BFS serialization and topology-diameter normalization of \cref{sec:representation}, we map every motion into a unified canonical space:
\begin{enumerate}
  \item Each animation is placed in a canonical coordinate frame with $y$-axis up and the initial root position at the origin.
  \item The initial facing direction is aligned with the positive $z$-axis, estimated as $\mathbf{f}=\mathrm{proj}_{xz}(\mathbf{e}_y \times \mathbf{v}_{\mathrm{hip}})$, where $\mathbf{v}_{\mathrm{hip}}$ is the left-to-right hip (or any symmetric-joint pair) direction.
  \item Joint rotations are expressed relative to the rest pose, yielding a unified kinematic basis across heterogeneous skeletons.
  \item We compute global statistics $(\mu_{\mathrm{global}},\sigma_{\mathrm{global}})\in\mathbb{R}^{D}$ over root-joint features and local statistics $(\mu_{\mathrm{local}},\sigma_{\mathrm{local}})\in\mathbb{R}^{D}$ over non-root features, and apply them to normalize all motions.
\end{enumerate}

\section{Experiments}
\label{sec:experiments}
We evaluate \themodel{} from five perspectives: presenting qualitative results across diverse rigs, comparing with skeleton-based motion generation methods, benchmarking against skeleton-free mesh animation baselines, demonstrating broader applications, and ablating our topology-aware design choices.

\subsection{Implementation Details}
\label{sec:implementation}
Our motion diffusion transformer consists of 8 blocks with hidden dimension 512. We use a frozen FLAN-T5~\cite{flan-t5} text encoder and train the model with AdamW~\cite{adamw} at a learning rate of $1\times10^{-4}$. To address the long-tailed topology distribution of \thedata{} and improve generalization, we adopt square-root-balanced sampling together with four kinematics-preserving on-the-fly augmentations: joint removal, joint addition, skeleton pooling~\cite{pool}, and bone-length perturbation. Training is performed with a global batch size of 256 on 8 NVIDIA H100 GPUs for one day, and inference runs at 50\,FPS.
\begin{figure*}[t]
  \centering
  \setlength{\belowcaptionskip}{\floatsep}
  \includegraphics[width=1.0\linewidth]{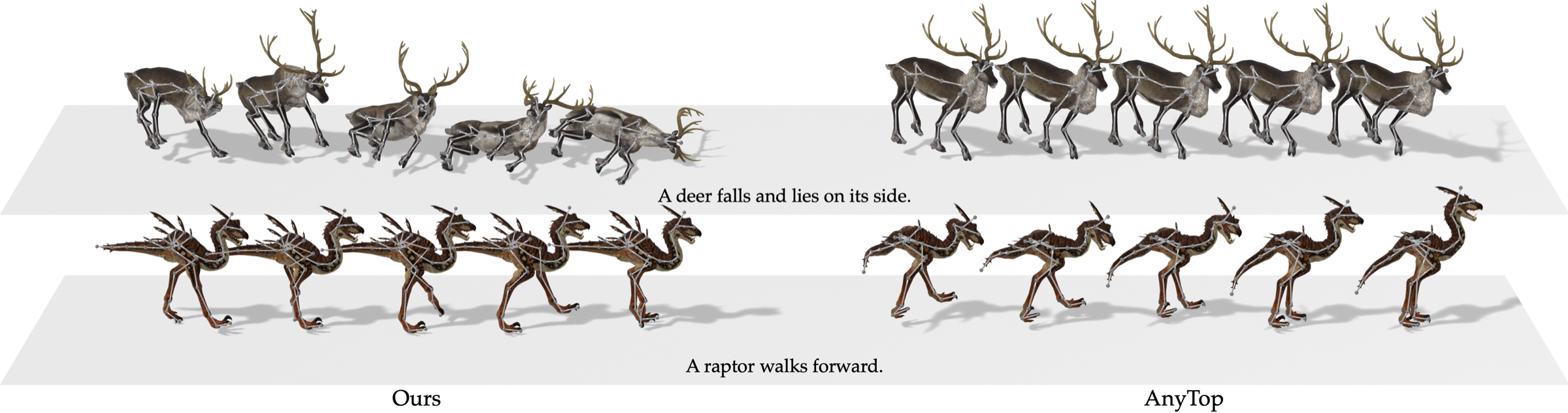}
  \Description{Side-by-side motion sequences comparing UniMate and AnyTop on unseen animal skeletons.}
      \caption{\textbf{Comparison with AnyTop on \underline{unseen} Truebones skeletons.} Ours (left) vs.\ AnyTop (right): our deer falls and lies on its side and our raptor walks forward as prompted, whereas AnyTop's deer never falls and its raptor remains nearly stationary.}
  \label{fig:anytop-comparison}

  \setlength{\belowcaptionskip}{0pt}
  \includegraphics[width=1.0\linewidth]{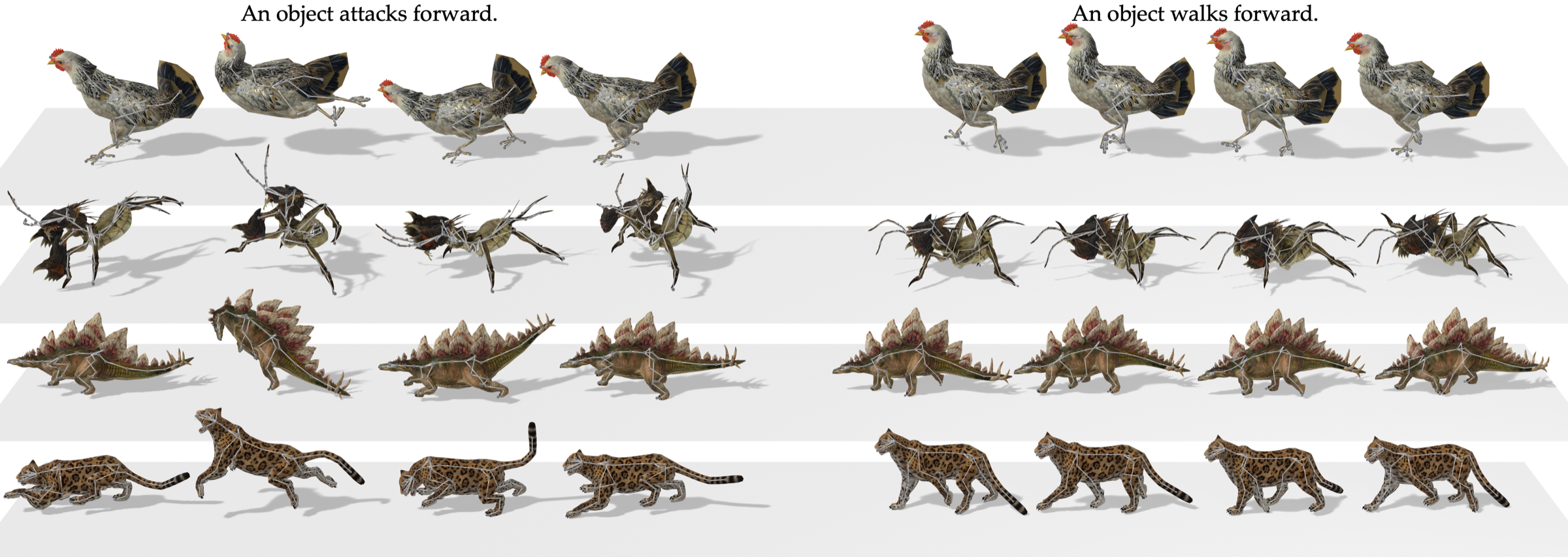}
  \Description{Two groups of animation sequences, one per text prompt. Each group shows four characters with different skeletal topologies---a bird, a many-legged insect, a stegosaurus, and a leopard---each animated over four frames with its skeleton overlaid on the mesh.}
      \caption{\textbf{Text-mediated motion transfer.} The source behavior is abstracted into a text prompt (shown atop each group); the same prompt then animates four target rigs of differing topology, conditioned directly on each target skeleton.}
  \label{fig:motion-transfer}
\end{figure*}
\begin{table*}[b]
  \centering
  \Description{Three quantitative comparisons reporting motion-generation quality and diversity, mesh-animation video metrics and runtime, and user-study ratings.}
  \renewcommand{\arraystretch}{1.15}
  \begin{minipage}[t]{0.25\textwidth}
  \centering
  \caption{\textbf{Comparison on motion generation.} Our model clearly outperforms AnyTop in both quality and diversity.}
  \label{tab:motion-generation}
  \scriptsize
  \setlength{\tabcolsep}{10pt}
  \begin{tabular}{@{}lcc@{}}
    \toprule
    \textbf{Method} & \textbf{FID}$\downarrow$ & \textbf{Div.}$\uparrow$ \\
    \midrule
    AnyTop~[\citeyear{anytop}] & 2.711 & 8.139 \\
    \textbf{\themodel\ (Ours)} & \textbf{0.757} & \textbf{9.200} \\
    \bottomrule
  \end{tabular}
\end{minipage}

  \hfill
  \begin{minipage}[t]{0.35\textwidth}
  \centering
  \caption{\textbf{Comparison on mesh animation.} Our method outperforms prior state-of-the-art approaches.}
  \label{tab:mesh-animation}
  \scriptsize
  \setlength{\tabcolsep}{3.25pt}
  \begin{tabular}{@{}l|cccc|c@{}}
    \toprule
    \textbf{Method} & OC\,$\uparrow$ & MS\,$\uparrow$ & DD\,$\uparrow$ & AQ\,$\uparrow$ & Time\,$\downarrow$ \\
    \midrule
    V2M4~[\citeyear{v2m4}] & 0.167 & 0.991 & 0.667 & 0.506 & 1.641h \\
    AnimateAnyMesh~[\citeyear{animateanymesh}] & 0.151 & \textbf{0.995} & 0.352 & 0.498 & 15.542s \\
    \textbf{\themodel\ (Ours)} & \textbf{0.186} & 0.993 & \textbf{0.833} & \textbf{0.544} & \textbf{1.214s} \\
    \bottomrule
  \end{tabular}
\end{minipage}

  \hfill
  \begin{minipage}[t]{0.38\textwidth}
  \centering
  \caption{\textbf{User study on mesh animation.} Our model achieves the best text--motion alignment and motion quality.}
  \label{tab:user-study}
  \scriptsize
  \setlength{\tabcolsep}{3.5pt}
  \begin{tabular}{@{}l|cccc|c@{}}
    \toprule
    \textbf{Method} & TA\,$\uparrow$ & MP\,$\uparrow$ & ME\,$\uparrow$ & SP\,$\uparrow$ & Avg.\,$\uparrow$ \\
    \midrule
    V2M4~[\citeyear{v2m4}] & 2.378 & 2.341 & 2.596 & 2.336 & 2.413 \\
    AnimateAnyMesh~[\citeyear{animateanymesh}] & 2.193 & 2.930 & 2.362 & 3.747 & 2.808 \\
    \textbf{\themodel\ (Ours)} & \textbf{4.617} & \textbf{4.568} & \textbf{4.646} & \textbf{4.630} & \textbf{4.615} \\
    \bottomrule
  \end{tabular}
\end{minipage}

\end{table*}

\subsection{Qualitative Results}
\label{sec:qualitative-results}
\cref{fig:qualitative-results,fig:cross-topology-animations} showcase animations generated by \themodel{} across heterogeneous rigs, spanning humanoids, quadrupeds, avians, insects, and articulated rigid objects: a single unified model produces prompt-faithful, temporally coherent motion while adapting to each skeleton's structure. \cref{fig:diverse-prompts,fig:diverse-motions} show that a single rig faithfully follows distinct prompts and yields diverse yet prompt-consistent samples under the same prompt, reflecting both the controllability and the generative diversity of the model, and \cref{fig:motion-long} demonstrates temporally coherent, drift-free long-horizon generation.
\begin{figure*}[t]
  \centering
  \includegraphics[width= 1.0\linewidth]{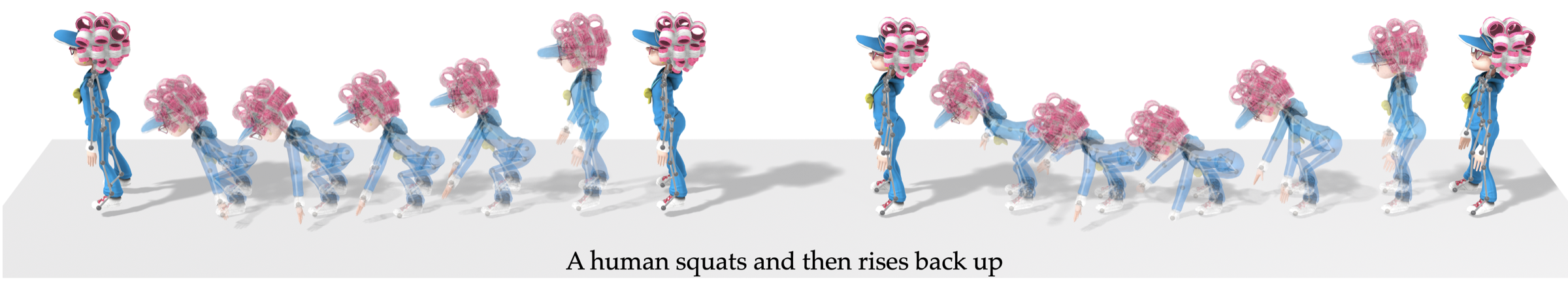}
  \Description{Motion in-betweening sequences showing fixed endpoint poses and synthesized intermediate poses.}
      \caption{\textbf{Motion in-betweening.} Given the start and end poses and a text prompt, \themodel{} synthesizes smooth and plausible intermediate motions; the opaque poses mark the given boundary constraints, while the translucent poses are the synthesized in-betweens.}
  \label{fig:motion-in-betweening}
  \includegraphics[width= 1.0\linewidth]{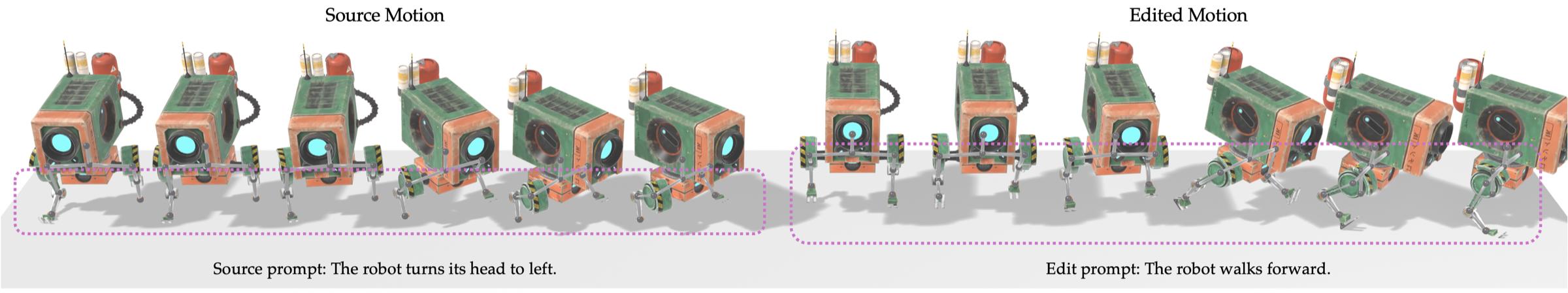}
  \Description{Text-guided motion editing examples with selected joints fixed and the resampled joints outlined by dashed boxes.}
      \caption{\textbf{Motion editing.} Given a source motion (left), a subset of joints stays fixed while the rest (dashed boxes) are resampled under a new prompt (right).}
  \label{fig:motion-editing}
  \includegraphics[width= 1.0\linewidth]{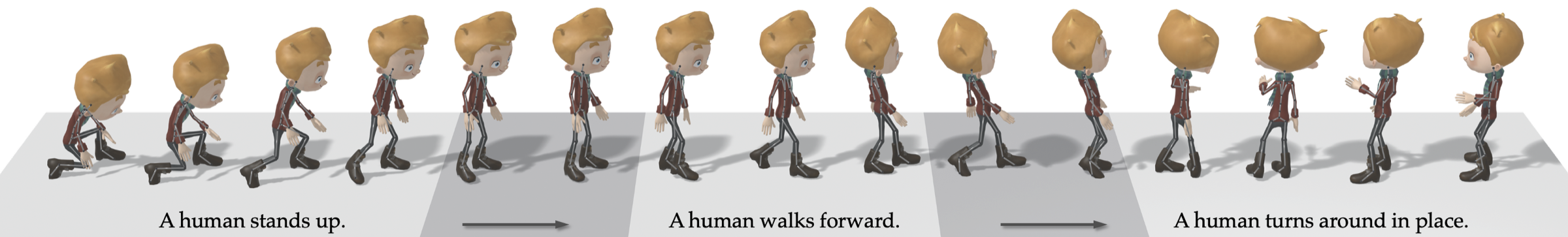}
  \Description{A motion-expansion sequence in which a character stands up, walks forward, and then turns around in place under three consecutive text prompts.}
      \caption{\textbf{Motion expansion.} Given sequential text prompts---here standing up, walking forward, then turning in place---\themodel{} extends a motion with smooth transitions; dark gray marks the boundary frames shared by consecutive segments.}
  \label{fig:motion-expansion}
\end{figure*}

\subsection{Comparison on Topology-Aware Motion Generation}
\label{sec:motion-generation-comparison}

\boldstartspace{Baselines.}
Our task is to generate a motion sequence conditioned on an input skeleton and a text prompt. To our knowledge, no existing baseline directly addresses this setting. The closest prior method is AnyTop~\cite{anytop}, which supports \emph{unconditional} motion generation for diverse skeletons on Truebones~\cite{truebones}. To adapt it to our setting and enable a fair comparison, we augment AnyTop with a cross-attention module for text conditioning, following MDM~\cite{mdm}. We train and evaluate both the extended AnyTop and our model on Truebones.

\boldstartspace{Metrics.}
We randomly hold out 7 skeleton types as unseen test skeletons, covering bipedal, quadrupedal, avian, marine, insectoid, and serpentine categories, for a total of 58 motion sequences and 5{,}861 frames. Specifically, (1) \emph{Motion quality} is measured by the Fréchet Inception Distance (FID) between extracted kinematic features~\cite{motion2motion,bailando} of generated and ground-truth motions, and (2) \emph{Diversity} is computed as the average pairwise joint distance among 5 generated samples. The held-out skeletons, prompts, and seed appear in \appendixref{app:heldout}.
\boldstartspace{Results.}
\themodel{} lowers FID from 2.711 to 0.757 and raises diversity from 8.139 to 9.200 (\cref{tab:motion-generation}), so fidelity is not bought with mode collapse, and the margin reflects the generalization of our topology-aware design to unseen skeletons. \cref{fig:anytop-comparison} qualitatively demonstrates that \themodel{} produces high-quality, prompt-faithful motion, whereas AnyTop either remains nearly static or ignores the prompt; \appendixref{app:anytop-additional} adds further morphologies and motions.
\subsection{Comparison on Text-Conditioned Mesh Animation}
\label{sec:mesh-animation-comparison}

\boldstartspace{Baselines.}
To further evaluate \themodel{} in a broader animation setting, we compare against two state-of-the-art \emph{skeleton-free} and \emph{vertex-wise} mesh animation baselines: (1) AnimateAnyMesh~\cite{animateanymesh}, a feed-forward, text-conditioned framework; and (2) V2M4~\cite{v2m4}, an optimization-based, monocular video-conditioned method. For a fair comparison, all methods are evaluated using the same text prompts. For V2M4, we use Wan2.2~\cite{wan} to generate the driving videos.

\boldstartspace{Metrics.}
Our benchmark consists of 18 randomly selected meshes spanning bipeds, quadrupeds, avians, marine life, insects, and general articulated objects. Following~\cite{animateanymesh,animax}, we render $512\times 512$ multi-view videos from fixed viewpoints and assess perceptual quality with VBench~\cite{vbench} along four axes: \emph{overall consistency} (OC), \emph{motion smoothness} (MS), \emph{dynamic degree} (DD), and \emph{aesthetic quality} (AQ). We also report the average generation time per mesh animation.
We further conduct a user study with 32 participants, each reviewing 12 test cases and rating each animation on a 5-point Likert scale ($1$ = very poor, $5$ = excellent) along \emph{text-to-motion agreement} (TA), \emph{motion plausibility} (MP), \emph{motion expressiveness} (ME), and \emph{shape preservation} (SP).

\boldstartspace{Results.}
In \cref{tab:mesh-animation,tab:user-study}, \themodel{} outperforms prior methods on most VBench metrics and achieves the highest scores across all user-study criteria, indicating stronger text--motion alignment and more expressive, plausible, and coherent motion; AnimateAnyMesh attains higher motion smoothness mainly due to near-static outputs with much lower dynamic degree and expressiveness, while V2M4 relies on separately generated driving videos and is substantially slower and less practical.

\subsection{More Applications}
\label{sec:applications}

All four applications below are zero-shot: they reuse the same pretrained model without fine-tuning or auxiliary networks, differing only in which motion tokens are held fixed during sampling.

\boldstartspace{Text-Mediated Motion Transfer.}
\themodel{} naturally supports cross-topology motion transfer, as shown in \cref{fig:motion-transfer}: a source motion is first abstracted into a language prompt, which then animates a target rig of different topology. Since generation is \emph{directly conditioned on the target skeleton}, no joint correspondence, exemplar alignment, or per-skeleton optimization is required.
\boldstartspace{Motion In-Betweening.}
\themodel{} further supports zero-shot motion in-betweening, as shown in \cref{fig:motion-in-betweening}. Given a target skeleton, a text prompt, and prescribed start and end poses, we perform sampling-time pose guidance by keeping the boundary-frame pose tokens fixed during the flow integration, while classifier-free text guidance enforces the motion semantics. This produces coherent transitions that satisfy the endpoint constraints while language specifies the transition's style and intent.
The same guidance extends beyond endpoints: keyframes of arbitrary number and position can be held fixed as a sampling-time mask, and \themodel{} infills coherent, prompt-consistent motion between them.

\boldstartspace{Motion Expansion.}
\themodel{} can extend an animation by chaining text prompts, as shown in \cref{fig:motion-expansion}: each new segment is generated with the preceding segment's final pose as a boundary condition, so the sequence grows smoothly while its semantics evolve.
Chaining complements the long-horizon variant of \cref{fig:motion-long}: the variant widens the temporal window of one sampling pass, while chaining composes arbitrarily many prompted segments.

\boldstartspace{Text-Guided Motion Editing.}
\cref{fig:motion-editing} showcases text-guided motion editing. Starting from an existing animation, we keep the unchanged parts fixed and resample selected joints under a new text prompt. This enables localized edits such as changing action intensity, modifying limb behavior, or altering motion direction without regenerating the entire sequence. Because the model jointly reasons over motion and topology, the edits remain temporally smooth and structurally consistent with the input rig.
Since only the selected joints are resampled, edits complete at interactive rates, supporting iterative prompt-driven refinement.

\subsection{Ablation Study}
\label{sec:ablation}
In \cref{tab:ablation}, we ablate the three topology-aware components of TADiT. Removing the graph-aware attention bias increases FID, showing the value of explicit structural relations in joint attention. Removing Spec-RoPE further degrades both fidelity and diversity, indicating weaker generalization to unseen skeletons. Removing the global topological conditioner causes the largest drop in motion quality; although diversity increases, the generated motions become less stable and often exhibit jitter.
\Appendixref{app:additional-results} complements these numbers with qualitative renderings of each ablated variant, along with further ablations of the data-processing pipeline.

\begin{table}[htb]
  \Description{Ablation results showing motion quality and diversity after removing each topology-aware component.}
  \centering
  \footnotesize
  \renewcommand{\arraystretch}{1.15}
  \caption{\textbf{Ablation.} Every topology-aware component improves quality.}
  \label{tab:ablation}
  \begin{tabular*}{0.75\columnwidth}{@{\extracolsep{\fill}}lcc@{}}
    \toprule
    \textbf{Method} & \textbf{FID}$\downarrow$ & \textbf{Diversity}$\uparrow$ \\
    \midrule
    w/o graph-aware attention bias & 0.773 & 8.799 \\
    w/o Spec-RoPE & 0.798 & 8.424 \\
    w/o global topological conditioner & 0.825 & \textbf{9.475} \\
    \midrule
    \textbf{Full Model} & \textbf{0.757} & 9.200 \\
    \bottomrule
  \end{tabular*}
\end{table}

\section{Limitations and Future Work}
\label{sec:limitations}

\begin{figure}[t]
  \centering
  \input{paper-figures/sliding-panel}
\end{figure}

\boldstartspace{Contact and Foot Sliding.}
\themodel{} can produce foot sliding, drift, hovering, or ground penetration in contact-rich motions (\cref{fig:sliding}), because it imposes no unified contact model: foot--ground contact is meaningful for bipeds and quadrupeds, but ill-defined for snakes, swimming fish, birds in flight, and many articulated objects. Future work could introduce morphology-aware contact objectives during constraint-guided sampling; where contacts are well defined, foot locking or IK post-processing can be applied to the predicted rig. \Appendixref{app:foot-contact} quantifies foot sliding on held-out legged skeletons and reports the effect of IK-based foot locking.

\begin{figure}[t]
  \centering
  \input{paper-figures/rare-topologies-panel}
\end{figure}

\boldstartspace{Rare Topologies and Motions.}
\themodel{} is less reliable on rare skeletal topologies and out-of-distribution motions, where results can become static, jittery, or semantically inaccurate (\cref{fig:rare-topologies}). This stems from the scarce, long-tailed 4D animation data, which favor humanoids and common locomotion. Future work could distill Internet-scale video priors to broaden topology and motion coverage while retaining \themodel{}'s topology-aware backbone.
Agentic asset-generation systems~\cite{articraft} offer a complementary data-side remedy: pairing automatically generated assets with scripted, simulated, or distilled motion could help densify the rare topology and motion regions where captured data is scarce.

\section{Conclusion}
\label{sec:conclusion}

We presented \themodel{}, a unified foundation model that synthesizes articulated motion for skeletons of arbitrary topology from a rigged 3D asset and a text prompt, with no test-time optimization or per-skeleton specialization. At its core, the Topology-Aware Diffusion Transformer couples motion and skeletal structure through a graph-aware attention bias, the Spec-RoPE spectral rotary position embedding, and a global topological conditioner, while \thedata{} provides large-scale motion supervision across thousands of heterogeneous skeletons. \themodel{} achieves state-of-the-art quality, generalization, and efficiency, and the same pretrained model supports motion transfer, in-betweening, expansion, and editing zero-shot. We hope \themodel{} and \thedata{} provide a foundation for scalable, controllable animation of arbitrary rigged assets, and that coupling them with video priors and agentic data generation will further broaden their coverage.

\bibliographystyle{ACM-Reference-Format}
\bibliography{references}

\clearpage
\begin{bibunit}[ACM-Reference-Format]
  \appendix

\lstset{
  basicstyle=\scriptsize\ttfamily,
  breaklines=true,
  breakatwhitespace=true,
  columns=fullflexible,
  keepspaces=true,
  frame=single,
  framesep=4pt,
  xleftmargin=4pt,
  xrightmargin=4pt,
  aboveskip=4pt,
  belowskip=4pt,
  upquote=true,
  showstringspaces=false,
  literate={°}{{\textdegree}}1 {—}{{--}}1 {→}{{->}}1 {∩}{{$\cap$}}1,
}

\section*{APPENDIX}
This appendix supplies material that the main paper defers for space. \cref{app:dataset} covers the dataset statistics, the LLM system prompts used for joint-name standardization, facing-direction joint-pair selection, and motion captioning, the four-view rendering camera setup, and the online skeletal augmentation pipeline. \cref{app:implementation} reports the architectural and training/inference hyperparameters for \themodel{} and the full expressions for the three training losses. \cref{app:spectral-rope-theory} gives a self-contained theoretical analysis of Spec-RoPE, covering its relative-coordinate structure, permutation equivariance, connection to standard RoPE, and effective-resistance interpretation. \cref{app:experiments} details our evaluation protocols, baselines, and the user-study setup, and \cref{app:additional-results} collects additional results: further AnyTop comparisons, direct comparisons with the skeleton-free baselines, qualitative ablations of the architecture and of the dataset-curation stages, and the foot-sliding analysis. Finally, \cref{app:limitations} expands the discussion of limitations and future work.

\paragraph{Code and data availability.}
The public project repository is available at \textcolor{ACMDarkBlue}{\url{https://github.com/Friedrich-M/UniMate}}. The repository currently serves as the permanent project entry point; upon publication, we will release the complete training and data-preprocessing code, pretrained model checkpoints, and the \thedata{} dataset at this URL.

\paragraph{Interactive demo.}
An interactive demo with more results is available at \textcolor{ACMDarkBlue}{\url{https://linzhanmou.com/unimate/interactive}}.

\section{Dataset Construction}
\label{app:dataset}

\subsection{Data Sources and Statistics}
\label{app:data-sources}
\thedata{} aggregates three complementary motion sources. Truebones~\cite{truebones} contributes 1{,}094 animal motion sequences spanning 74 distinct skeletons. Mixamo~\cite{mixamo} adds 2{,}425 high-quality human sequences. From Objaverse-XL~\cite{objaverse,objaversexl} we filter and deduplicate 6{,}965 articulated assets with both rigging~\cite{puppeteer} and animation~\cite{diffusion4d} annotations, yielding 9{,}487 valid action sequences. In total, \thedata{} comprises 13{,}006 sequences and 2{,}140{,}232 frames over thousands of unique skeletons. \Cref{alg:uniml3d-data-processing} lists the per-stage operations of the curation pipeline summarized in the main paper.

\begin{figure}[h]
  \centering
  \includegraphics[width= 1.0\linewidth]{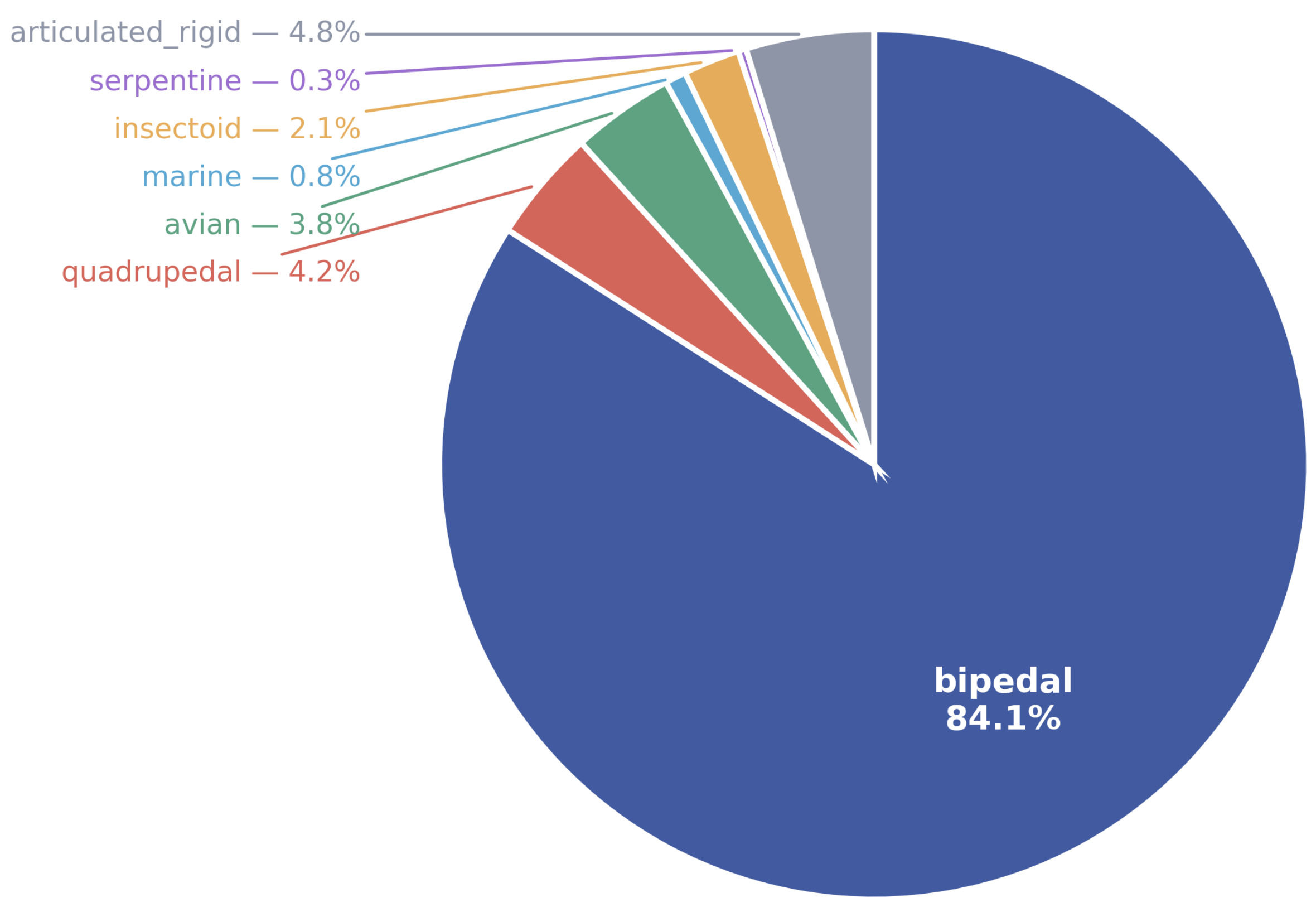}
  \Description{Pie chart of the share of UniML3D motion sequences per skeletal morphology category, dominated by bipedal characters.}
  \caption{\textbf{Morphology distribution of \thedata.} Share of motion sequences per skeletal morphology; bipedal characters dominate, while serpentine and marine rigs are rare.}
  \label{fig:category-statistics}
\end{figure}

\begin{figure}[h]
  \centering
  \includegraphics[width= 1.0\linewidth]{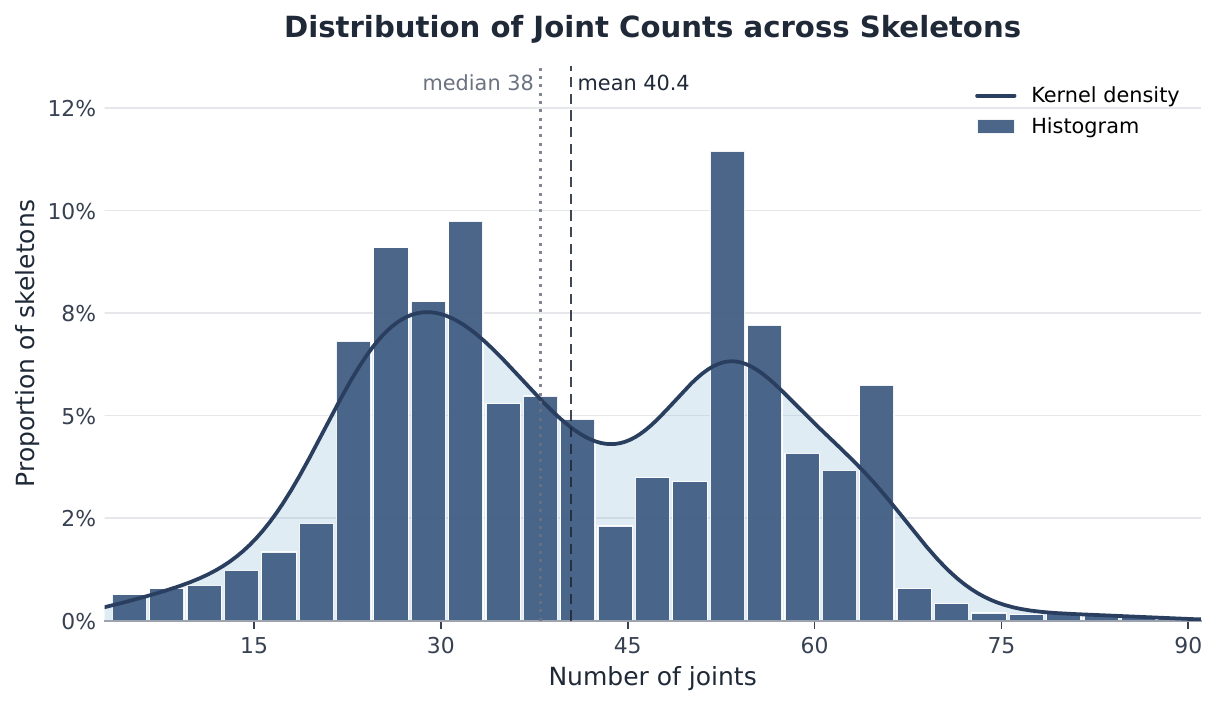}
  \Description{Histogram of joint counts across skeletons retained in UniML3D.}
      \caption{\textbf{Joint-count distribution across skeletons.} Histogram and kernel density of the number of joints per retained skeleton in \thedata{} (median 38, mean 40.4); the bimodal spread shows that the dataset covers rigs of widely varying complexity.}
  \label{fig:joint-count-distribution}
\end{figure}

\Cref{fig:category-statistics,fig:joint-count-distribution} visualize the distribution of \thedata{} across morphology categories and joint counts. The category distribution is heavily long-tailed: bipedal characters dominate at $84.1\%$, while serpentine ($0.3\%$) and marine ($0.8\%$) categories are sparsely populated, which motivates the square-root-balanced sampler defined in \cref{app:augmentation}. The joint-count distribution further shows that the dataset spans skeletons of widely varying complexity, with concentrations around the typical humanoid rigging conventions.

\begin{algorithm*}[t]
  \caption{\thedata{} data processing. The three stages of \cref{sec:dataset} act at three granularities: skeleton-level steps run once per rig, clip-level steps once per clip, and feature normalization once over the whole set. A clip either passes every check or is discarded.}
  \label{alg:uniml3d-data-processing}
  \begin{algorithmic}[1]
    \Require rigged assets from Truebones, Mixamo, and Objaverse-XL, each a skeleton $\mathcal{S}$ with a set of animation clips $\mathcal{X}$
    \Ensure the training set of canonical skeleton--motion--caption samples $(\mathcal{S}, x, c)$
    \ForAll{rigged assets $(\mathcal{S}, \mathcal{X})$ in the source pool}
      \Statex \emph{Skeleton-based filtering, skeleton level (\cref{sec:dataset})}
      \State $\mathcal{S} \gets$ the tree with the largest cumulative skinning weight; realign a spurious root onto the semantic root via forward kinematics; prune zero-weight helper joints
      \Statex \emph{Semantic annotation, skeleton level (\cref{app:joint-name-standardization,app:facing-pair})}
      \State $\mathcal{N}_{\mathcal{S}} \gets$ joint names standardized to the anatomical vocabulary by an LLM
      \State $(j_{\mathrm{L}}, j_{\mathrm{R}}) \gets$ the symmetric joint pair defining the lateral axis, selected by an LLM from $\mathcal{N}_{\mathcal{S}}$
      \ForAll{clips $x \in \mathcal{X}$}
        \Statex \emph{Skeleton-based filtering, clip level (\cref{sec:dataset})}
        \If{$x$ is static \textbf{or} physically implausible under the criteria of \cref{sec:dataset}}
          \State \textbf{discard} $x$ and continue with the next clip
        \EndIf
        \Statex \emph{Semantic annotation, clip level (\cref{app:rendering,app:motion-captioning})}
        \State resample $x$ to 30\,FPS; render four synchronized views of $x$ and of the rest pose; $c \gets$ caption of the rendering by a multimodal LLM
        \Statex \emph{Motion canonicalization, clip level (\cref{sec:representation,sec:dataset})}
        \State serialize $\mathcal{S}$ in breadth-first order from the root; scale $\mathcal{S}$ and $x$ by the topology diameter $d_{\mathrm{topo}}$
        \State place $x$ in a $y$-up frame with the initial root at the origin; align the initial facing direction $\mathbf{f}$, derived from the $j_{\mathrm{L}} \to j_{\mathrm{R}}$ direction as in \cref{sec:dataset}, with $+z$
        \State express joint rotations relative to the rest pose; assemble the per-joint features $\mathbf{m}_j^t = [\mathbf{p}_j^t, \mathbf{r}_j^t, \mathbf{v}_j^t]$; add $(\mathcal{S}, x, c)$ to the training set
      \EndFor
    \EndFor
    \Statex \emph{Motion canonicalization, dataset level (\cref{sec:dataset})}
    \State normalize root features with $(\mu_{\mathrm{global}}, \sigma_{\mathrm{global}})$ and non-root features with $(\mu_{\mathrm{local}}, \sigma_{\mathrm{local}})$, both computed over the whole training set
  \end{algorithmic}
\end{algorithm*}

\subsection{Joint Name Standardization}
\label{app:joint-name-standardization}
\begin{sloppypar}
Raw joint labels in our source datasets are highly inconsistent: they mix DCC-tool namespaces (\texttt{mixamorig:}, \texttt{QuickRigCharacter\_}, \texttt{Bip01\_}), Maya/Blender suffixes (\texttt{\_jnt}, \texttt{.001}), chain indices (\texttt{Spine1}, \texttt{Thumb1}), 3ds Max Biped numeric finger codes (\texttt{Finger01}, \texttt{Finger21}), Japanese romaji roots used by some animal rigs (\texttt{momo}, \texttt{munabire}), and idiosyncratic per-asset placeholders. Objaverse-XL exports additionally append a per-joint index to every name, sometimes on top of the original chain index (\texttt{Hip\_01\_41}, \texttt{Index.R.001\_013\_6}). To make the per-joint name embeddings used by \themodel{} comparable across rigs, we standardize every joint label to a fixed anatomical vocabulary using DeepSeek-V4-Flash~\cite{deepseek-v4} with the deterministic system prompt below. Each rig is processed as a whole, so the model can resolve a joint's role from its neighbors in the hierarchy, and every input joint must map to exactly one label. Responses that violate this one-to-one correspondence are rejected and re-queried, and rigs that repeatedly fail fall back to a deterministic rule-based cleaner over the same vocabulary. An optional refinement pass with GPT-5~\cite{gpt5} then re-examines each label alongside its raw name and corrects residual errors.
\end{sloppypar}

\begin{lstlisting}
Standardize 3D rig joint names to canonical anatomical labels. Inputs come from Mixamo, Maya, Blender, Unreal, Truebones and custom rigs. Focus on semantic meaning, not surface syntax.

OUTPUT: a single JSON array of strings, same length and order as the input. No prose, no fences, no extra text. One input entry -> one output entry; never dedupe, merge, skip or reorder, even when neighbours produce identical labels.

CLEAN each name by removing rig noise and extracting meaning:
1. Digits and Blender '.NNN' counters are meaningless — rig-internal bookkeeping (chain index, mirror id, duplicate counter). Ignore them when matching, and never include a digit, underscore, or dot in the final label. 'Spine', 'Spine1', 'Spine_02', 'Spine.003' all map to 'Spine'. Finger-chain segments ('Thumb1/Thumb2/Thumb3') all map to 'Thumb Finger'. The Objaverse export pipeline also stamps a trailing global index on EVERY name ('_NN' or '_0NN': '_01', '_010', '_063'); indices can STACK ('Hip_01_41', 'Head_1_016', 'Index.R.001_013_6', 'Bone.001_01', 'Spine_1_013') — strip ALL of them, in any position. Editor decorations are noise too: strip '(mirrored)' anywhere and a trailing '.x' center marker ('spine_01.x' -> 'Spine').
2. Drop any leading '<Word>:' namespace (case-insensitive, trailing digits in the namespace OK): mixamorig:, Mixamorig1:, Mutant:, Sif:. The same words are noise without the colon too — leading 'mixamorig_'/'Mixamorig'/'Character'/'Rig'/'QuickRigCharacter_' segments are dropped, never echoed into the label. The same goes for embedded ASSET/CHARACTER names and their decorations — 'rp_karl_animated_006_warmingUp_spine_01' -> 'Spine', 'CMan0205-M4-CS_Hips L Finger0' -> 'Left Thumb Finger': keep only the anatomical tokens, drop every name-like or counter token.
3. Drop rig prefixes (match before ignoring digits so 'Bip01_' still strips cleanly): any Bip<digits> container with any separator (Bip01_, Bip002 , Bip01-, and separator-free Bip001LFinger0), BN_Bip01_, BN_, Bn_, NPC_, jt_, Elk, Sabrecat_, QuickRigCharacter_, Bind_, Skeleton_, Root_, DEF-, def_.
4. Drop Maya suffixes: _jnt, _jt, _Jt, _JNT, _joint, _bone, _bn, _C.
5. Extract side as explicit 'Left '/'Right ' prefix. Recognise:
   prefix  L_ / R_, Lt_ / Rt_, Left_ / Right_, Left<UpperWord> / Right<UpperWord> (LeftHand, RightArm), L<UpperLetter> / R<UpperLetter> (LArm, RHand) — but NOT when followed by a lowercase letter (Lower, Ribcage).
   3ds Max Biped  space-separated single-letter side: 'Bip001 L UpperArm' -> Left Upper Arm, 'Bip001 R Thigh' -> Right Thigh. Token boundaries are spaces, not underscores.
   suffix  _L / _R / _l / _r, .L / .R, Japanese trailing L/R.
   PRECEDENCE: a trailing .L/.R/_L/_R token OVERRIDES a leading side word — Blender's symmetrize renames only the suffix, leaving the prefix text stale: 'mixamorig:LeftShoulder.R' -> 'Right Shoulder', 'r_toe.L' -> 'Left Toe'.
   quadruped F_/B_ = Front/Back (e.g. 'F_R_Shoulder' -> 'Right Front Shoulder', 'B_L_Foot' -> 'Left Back Foot').
6. Common body-part roots — translate case-insensitively, accepting Unreal snake_case AND CamelCase short-forms: pelvis/spine/neck/head/jaw/eye -> Pelvis/Spine/Neck/Head/Jaw/Eye; clavicle/collar -> Shoulder; upperarm / UpperArm / UpArm -> Upper Arm; lowerarm / LowerArm / LowArm / ForeArm -> Forearm; hand -> Hand; thigh / upleg / UpLeg -> Thigh; calf / lowleg / LowLeg / lowerleg -> Shin; upperleg -> Thigh; toes -> Toe; in a mixamo chain (UpLeg -> Leg -> Foot) the mid-bone 'Leg' is the shin -> Shin; foot -> Foot; toebase / Toe0 / toe -> Toe; ball -> Toe; eyelid -> Eyelid. Finger roots index/middle/ring/pinky/thumb -> '<Root> Finger'. *_twist -> '<Root> Twist'.
6b. 3ds Max Biped fingers use numeric codes: Finger0*=Thumb, Finger1*=Index, Finger2*=Middle, Finger3*=Ring, Finger4*=Pinky. Any trailing digits after that code are chain position — ignore. 'Bip001 L Finger0' and 'Bip001 L Finger01' both -> 'Left Thumb Finger'; 'Bip001 R Finger21' -> 'Right Middle Finger'; 'Bip001 R Toe0' -> 'Right Toe'.
6c. Mocap-segmented fingers are 1-BASED and carry a segment word: Finger1..Finger5 + Metacarpal/Proximal/Medial/Distal/Tip, with Finger1=Thumb ... Finger5=Pinky. 'LeftFinger1Metacarpal' -> 'Left Thumb Finger'; the Tip segment -> '<Root> Finger End'. Rule 6b's 0-based codes apply only to BARE Finger<digit> names with no segment word. Segment words after a NAMED finger ('IndexDistal', 'thumb_proximal_l', 'RingIntermediate') are likewise chain position — drop them: 'IndexDistal' -> 'Index Finger'.
7. Other direction words: Top->Upper, Low->Lower ('Topjaw'->'Upper Jaw'). 'HeadTop_End' / '*_End' / '*Nub' -> '<Root> End'. Animal 'Hair*/Mane*' -> 'Mane'; humanoid accessories 'Ponytail*/Cape*/Cloth*/Skirt*' -> 'Appendage'.
8. Placeholders -> 'Bone': Bone, joint, Xtra*, MagicEffectsNode, and any token with no clear anatomy (meshok, Capuche, ...). Do NOT fabricate body parts. EXCEPTION: a name that is ONLY digits, optionally with a leading underscore ('_00', '12'), is copied through UNCHANGED — it marks a rig with unnamed bones.
9. Japanese roots (Alligator/Pirrana/Tukan):
   body   momo=Thigh, hiza=Knee, ashi=Foot, hiji=Elbow, te=Hand, kata=Shoulder, mune=Chest, hara=Abdomen, koshi/kosi=Hips, kubi=Neck, atama/kao=Head, ago=Jaw
   tail   sippo/shippo=Tail, o=Tail
   fish   munabire=Pectoral Fin, harabire=Pelvic Fin, sebire=Dorsal Fin, obire=Caudal Fin, shiribire=Anal Fin, era=Gill
   Trailing L/R on any of these -> Left/Right prefix.
10. Output Title Case, single spaces. Prefer the CANONICAL vocabulary; if nothing fits, use 'Bone'.

ALIASES & TYPOS (map to the canonical term): Spline=Spine, Scull=Skull, Nek=Neck, Tai=Tail, Tone/Thouge/Tunge=Tongue, Eyeleds=Eyelid, HorseLink=Fetlock, LargeCannon=Cannon, PhalanxPrima=Pastern, PhalangesManus=Phalanges, Foreleg=Front Leg, Hindleg=Hind Leg, Digit=Finger, Hair=Mane, Little=Pinky (finger), locator/Trajectory/Cog=Root, Clavicle/Collarbone=Shoulder. Species terms: insect 'Clip'/'Shall'=Mandible, 'Pliers'/'Piers'=Pincer, cricket 'Feeler'=Antenna but fish 'Feelers'=Barbel, bird 'ponitail'=Crest.

CANONICAL VOCABULARY (use these exact terms verbatim):
  Core: Pelvis, Hips, Spine, Ribcage, Neck, Head, Skull, Skull Base, Head End, Body, Upper Body, Lower Body, Chest, Abdomen, Waist, Collar, Hip, Belly, Root, Center
  Arm : Shoulder, Scapula, Arm, Upper Arm, Forearm, Elbow, Wrist, Hand, Palm
  Leg : Thigh, Shin, Leg, Knee, Ankle, Foot, Heel, Toe, Paw, Hoof, Fetlock, Cannon, Metacarpus, Phalanges, Pastern
  Finger: Finger, Thumb Finger, Index Finger, Middle Finger, Ring Finger, Pinky Finger
  Head: Jaw, Upper Jaw, Lower Jaw, Tongue, Ear, Eye, Eyeball, Eyebrow, Eyelid, Mouth, Lip, Upper Lip, Lower Lip, Nose, Muzzle, Chin, Cheek
  Appendage: Tail, Wing, Feather, Antenna, Barbel, Tentacle, Claw, Hand Claw, Fang, Mandible, Large Mandible, Lower Mandible, Pincer, Stinger, Appendage
  Fins: Fin, Pectoral Fin, Pelvic Fin, Dorsal Fin, Caudal Fin, Anal Fin, Gill
  Coat/Equipment: Mane, Fur, Whisker, Shell, Dorsal Plate, Crest, Horn, Reins, Halter
  Quadruped: Front Leg, Middle Leg, Hind Leg, Front Paw, Back Paw, Front Hoof, Rear Hoof, Front Shoulder, Back Hip
  Physics: Twist, Upper Arm Twist, Forearm Twist, Thigh Twist, Shin Twist, Thigh Muscle, Neck Muscle, Tail Twist, Jiggle, Handle, IK Chain, Bone
Side goes first ('Left Upper Arm', 'Right Pinky Finger'); quadruped markers sit between side and part ('Right Front Shoulder'). COMPOSED labels are also valid: '<part> End' for chain tips/Nub bones, 'Inner/Middle/Outer <part>' (raptor toes, claws, fingers), 'Upper/Lower Eyelid', 'Upper/Lower Left/Right/Front Lip', 'Wrist Back', 'Elbow Back', 'Front/Middle/Hind Leg End'.

EXAMPLES (one per rig family):
  mixamorig:LeftUpLeg         -> Left Thigh
  Mutant:RightHandThumb1      -> Right Thumb Finger
  Sif:calf_twist_01_r         -> Right Shin Twist
  index_01_l                  -> Left Index Finger
  Bip01_L_Thigh               -> Left Thigh
  BN_Bip01_R_Forearm_03       -> Right Forearm
  NPC_L_Finger02              -> Left Finger
  Elk_RearHoof_L              -> Left Rear Hoof
  jt_FrontLeg1_R_C            -> Right Front Leg
  Lt_Thumb1_jt                -> Left Thumb Finger
  R_toeBase_jnt               -> Right Toe
  Eye.R.001                   -> Right Eye
  mixamorig:LeftShoulder.R    -> Right Shoulder
  LeftHandIndex1              -> Left Index Finger
  mixamorig:RightLeg          -> Right Shin
  LeftFinger2Distal           -> Left Index Finger
  BN_Spline_03                -> Spine
  Bone.001                    -> Bone
  F_R_Shoulder                -> Right Front Shoulder
  B_L_Foot                    -> Left Back Foot
  Topjaw                      -> Upper Jaw
  momoR                       -> Right Thigh
  munabireL                   -> Left Pectoral Fin
  joint12 / Xtra01 / meshok   -> Bone
OBJAVERSE (trailing global '_NN', often stacked with a chain index):
  mixamorig:LeftUpLeg_056     -> Left Thigh
  mixamorig:LeftHandThumb1_012 -> Left Thumb Finger
  QuickRigCharacter_LeftForeArm_014 -> Left Forearm
  Bip001 L UpperArm_07        -> Left Upper Arm
  Bip001 L Finger01_011       -> Left Thumb Finger
  Bip001 R Finger21_036       -> Right Middle Finger
  Bip001 R Toe0_054           -> Right Toe
  UpArm.R_010_15              -> Right Upper Arm
  LowLeg.L_038_39             -> Left Shin
  Hip_01_41                   -> Pelvis
  Index.R.001_013_6           -> Right Index Finger
  Rt_Eyelid_jt_08             -> Right Eyelid
  Skeleton_Root_02            -> Root
  joint1_2 / Bone.001_01      -> Bone

BATCH EXAMPLE (notice identical outputs are preserved, not merged):
  input  : ["Spine", "Spine1", "Spine2", "Spine3", "Neck", "Tail_01", "Tail_02", "Tail_03"]
  output : ["Spine", "Spine", "Spine", "Spine", "Neck", "Tail", "Tail", "Tail"]
  WRONG  : ["Spine", "Neck", "Tail"]  (8 inputs must yield 8 outputs — deduping is a failure)
\end{lstlisting}

\subsection{Facing-Direction Joint-Pair Selection}
\label{app:facing-pair}
The canonicalization stage of the main paper aligns each clip's initial facing direction with the positive $z$-axis using the left-to-right direction of a bilaterally symmetric joint pair. Which pair plays that role differs from rig to rig (the thighs of a humanoid, the front shoulders of a quadruped, the pectoral fins of a fish), so we select it per rig with DeepSeek-V4-Flash, reasoning over the standardized joint names of \cref{app:joint-name-standardization}. A deterministic rule-based resolver first proposes a pair from a fixed priority order of body parts; the model then reviews the candidate joints (right-side, left-side, head, and tail) together with that proposal and follows a three-step decision. It pairs the right and left joints that mirror the highest-priority body part present on both sides; if no mirrored pair exists but head and tail joints do, as for serpentine rigs, it returns the head and tail chain endpoints as a longitudinal body axis instead; otherwise it returns no pair, which marks rigs without a meaningful lateral axis, such as vehicles and props. Invalid responses fall back to the rule-based proposal, and an optional GPT-5 refinement pass revisits the rigs left without a pair. The system prompt follows.

\begin{lstlisting}
# TASK
Pick the joint pair that defines one 3D rig's lateral facing axis. Rigs come from Objaverse (Mixamo, 3ds Max Biped, Maya QuickRig, Unreal mannequin, CAT, custom) and Truebones animals. All clean names are already normalised to a canonical vocabulary — reason on CLEAN names; RAW names are only for verbatim copy-back.

# INPUT (user message)
Four pre-filtered buckets (any may be empty):
  RIGHT : rows whose clean name starts with 'Right '
  LEFT  : rows whose clean name starts with 'Left '
  HEAD  : rows whose clean name (minus any trailing ' N') is one of
          {Head, Skull, Skull Base, Head End, Jaw, Upper Jaw,
           Lower Jaw, Tongue, Muzzle, Nose, Chin}
  TAIL  : rows whose clean name (minus any trailing ' N') is 'Tail'
          or 'Tail Twist'
Plus a 'Rule-based hint' JSON object — the deterministic resolver's pick. Use it as a sanity check, not ground truth (see HINT below).

# ALGORITHM
Execute the steps in order. Do not skip ahead.
STEP 1. Compute OVERLAP.
  - For each row in RIGHT, suffix = clean_name minus 'Right ' minus any trailing ' <digits>'. Collect into set R_SUFFIXES.
  - For each row in LEFT,  suffix = clean_name minus 'Left ' minus any trailing ' <digits>'. Collect into set L_SUFFIXES.
  - OVERLAP = R_SUFFIXES ∩ L_SUFFIXES.
STEP 2. If OVERLAP is non-empty — APPLY RULE A AND RETURN.
  a. Pick suffix `s` = first element of OVERLAP that appears in PRIORITY (see below). If none appear, pick the alphabetically first element of OVERLAP.
  b. R_ROWS = RIGHT rows whose suffix == s; L_ROWS = LEFT rows whose suffix == s.
  c. If len(R_ROWS)==1 and len(L_ROWS)==1, pair them.
     Otherwise (chain duplicates, e.g. scorpion legs) match by DIGIT SIGNATURE: the tuple of ALL digit groups in the RAW name ('Bip01_R_Thigh_4' -> (01,4)). Tier 1 — pair rows whose full signatures are equal. Tier 2 — no tier-1 match: drop the LAST group (the Objaverse per-joint global index: 'Bip01_R_Thigh_1_053' and 'Bip01_L_Thigh_1_054' both reduce to (01,1)) and pair on the rest. If neither tier matches, take R_ROWS[0] and L_ROWS[0].
  d. Output: source = s.lower(); body_axis = false.
  e. DO NOT consider HEAD or TAIL. Even if HEAD+TAIL look perfect, rule A wins whenever OVERLAP is non-empty.
STEP 3. OVERLAP is empty. If HEAD is non-empty AND TAIL is non-empty — APPLY RULE B AND RETURN.
  - r_hip = LAST HEAD row; l_hip = LAST TAIL row (chain endpoints).
  - source = 'body_axis'; body_axis = true.
STEP 4. Otherwise — APPLY RULE C (empty).
  - Both hips are {raw: '', clean: ''}. source = 'empty'; body_axis = false.
  - Vehicles, props, and abstract rigs land here. NEVER pair an unrelated Right/Left entry (e.g. lone 'Right Eye' without left mate) just to avoid emitting empty.

# PRIORITY (highest → lowest; first match in OVERLAP wins)
Hip-level  : thigh, shoulder, front shoulder, back hip, hip, scapula
Whole-limb : upper arm, arm, front leg, hind leg, middle leg, back leg, wing, leg
Aquatic    : pectoral fin, pelvic fin, fin, gill
Arthropod  : pincer, mandible, large mandible, lower mandible, stinger, claw, hand claw, antenna
Mid-limb   : forearm, shin, knee, elbow, ankle, wrist
Extremity  : hand, palm, foot, heel, front paw, back paw, paw, front hoof, rear hoof, hoof, fetlock, cannon, metacarpus, pastern, toe
Fingers    : thumb finger, index finger, middle finger, ring finger, pinky finger, finger, neck
Weak       : eye, eyeball, eyelid, eyebrow, ear, horn, cheek, whisker, fang, barbel, tentacle, feather
Last       : tail
Suffix comparison is case-insensitive but uses the clean-name form ('Upper Arm', 'Front Shoulder', 'Pectoral Fin'). Read the list above LITERALLY, left to right, top to bottom — it is exactly the order the rule-based resolver uses, so do not re-rank quadruped markers ('Front Shoulder', 'Back Hip', 'Front Leg') against their plain counterparts ('Shoulder', 'Hip', 'Leg'). Note in particular that 'shoulder' comes BEFORE 'front shoulder', while 'front leg' comes BEFORE 'leg'.

# HINT
The hint is the rule-based resolver's output. It is usually right but can err on edge cases:
  - Hint says 'body_axis' or 'empty' yet OVERLAP is non-empty → hint is wrong, apply STEP 2 (rule A wins).
  - Hint swapped sides (Right/Left in wrong fields) → fix.
  - Hint chose a lower-priority suffix than STEP 2a finds → override with the higher-priority one.
  - Hint paired mismatched chain indices on a multi-leg rig → fix via STEP 2c raw-suffix match.
  - Your algorithm yields the same answer as the hint → return the hint verbatim.

# OUTPUT
Exactly one JSON object, no prose, no markdown fences, no comments:
  {"r_hip":    {"raw": "<exact>", "clean": "<exact>"},
   "l_hip":    {"raw": "<exact>", "clean": "<exact>"},
   "source":   "<lowercase suffix or 'body_axis' or 'empty'>",
   "body_axis": <true|false>}

# INVARIANTS (verify before emitting)
  I1. r_hip.raw != l_hip.raw, UNLESS both are '' (empty case).
  I2. r_hip.raw and l_hip.raw each appear verbatim in the rig's input rows (copied character-for-character).
  I3. r_hip always holds the Right (or head) side; l_hip always holds the Left (or tail) side. Never swap.
  I4. body_axis == true  IFF  source == 'body_axis'.
  I5. source is lowercase. If rule A fired, source is the suffix lowercased with single spaces (e.g. 'thigh', 'front shoulder', 'pectoral fin').
  I6. Unless body_axis, r_hip and l_hip mirror the SAME part: their clean names minus the side prefix and any trailing digits are identical. Never pair different parts ('Right Thigh' with 'Left Shoulder' is invalid even though the sides are correct).

# WORKED EXAMPLES
## Ex1 — Mixamo humanoid (rule A, hip-level pick)
RIGHT has 'Right Thigh' + 'Right Shoulder'; LEFT has the mirrors. OVERLAP = {Thigh, Shoulder}. Thigh outranks Shoulder → pick Thigh.
  {"r_hip": {"raw": "mixamorig:RightUpLeg_056", "clean": "Right Thigh"},
   "l_hip": {"raw": "mixamorig:LeftUpLeg_056",  "clean": "Left Thigh"},
   "source": "thigh", "body_axis": false}
## Ex2 — Objaverse quadruped (Front Shoulder beats Back Hip)
No Thigh pair; OVERLAP = {Front Shoulder, Back Hip}. Front Shoulder outranks Back Hip.
  {"r_hip": {"raw": "F_R_Shoulder_012", "clean": "Right Front Shoulder"},
   "l_hip": {"raw": "F_L_Shoulder_012", "clean": "Left Front Shoulder"},
   "source": "front shoulder", "body_axis": false}
## Ex3 — Scorpion multi-leg (STEP 2c raw-suffix match)
RIGHT has three 'Right Thigh' rows (Bip01_R_Thigh_1/_2/_4); LEFT the mirrors. OVERLAP = {Thigh}. Digit signatures: (01,1) on both sides -> tier-1 match.
  {"r_hip": {"raw": "Bip01_R_Thigh_1", "clean": "Right Thigh"},
   "l_hip": {"raw": "Bip01_L_Thigh_1", "clean": "Left Thigh"},
   "source": "thigh", "body_axis": false}
## Ex4 — OVERRIDE a mistaken body_axis hint
RIGHT has 'Right Thigh'; LEFT has 'Left Thigh'; HEAD has 'Tongue'; TAIL has 'Tail'. Hint = body_axis Tongue+Tail. OVERLAP = {Thigh} is non-empty → STEP 2 wins, hint is wrong.
  {"r_hip": {"raw": "RightUpLeg_033", "clean": "Right Thigh"},
   "l_hip": {"raw": "LeftUpLeg_028",  "clean": "Left Thigh"},
   "source": "thigh", "body_axis": false}
## Ex5 — Snake / serpent (rule B fires)
RIGHT and LEFT are empty; HEAD has 'Head'; TAIL has 'Tail_30'. STEP 3 applies.
  {"r_hip": {"raw": "Head",    "clean": "Head"},
   "l_hip": {"raw": "Tail_30", "clean": "Tail"},
   "source": "body_axis", "body_axis": true}
## Ex6 — Vehicle / prop (rule C fires)
All sections empty, or only a lone 'Right Eye' with no left mate.
  {"r_hip": {"raw": "", "clean": ""},
   "l_hip": {"raw": "", "clean": ""},
   "source": "empty", "body_axis": false}
\end{lstlisting}

\subsection{Motion Rendering}
\label{app:rendering}
For caption supervision and qualitative inspection, every retained clip is rendered to a four-view synchronized video with an automated Blender pipeline that follows the camera convention of AnimaX~\cite{animax} and DIMO~\cite{dimo}. For each clip we additionally render the rest-pose asset under the same camera setup, providing the captioner with a static reference frame against which articulated motion can be read out. Four cameras are placed at fixed elevation $0^\circ$ and orthogonal azimuths $a \in \{0^\circ, 90^\circ, 180^\circ, 270^\circ\}$ on a sphere of radius $2$\,m around the rig, with a fixed field of view of $33.9^\circ$ and a pinhole projection. All motions are resampled to $30$\,FPS prior to rendering so that clip-level temporal statistics, frame-rate-dependent motion descriptors, and the captioning model's perceived motion speed are comparable across sources.

\subsection{Motion Captioning}
\label{app:motion-captioning}
We caption each animation clip by querying a multimodal LLM (Qwen3.5-9B~\cite{qwen35}) on the four-view rendering described in \cref{app:rendering}, paired with a task-specific system prompt. Three prompts target the human (Mixamo), generic (Objaverse-XL), and animal (Truebones) subsets of \thedata. All three share the same structure (camera description, observation protocol, caption rules, and style examples) and differ in the subject term, heading cues, body-part vocabulary, and dataset-specific rules; for Mixamo and Truebones, the source catalogue's motion label is supplied as a hint that the model may use to disambiguate but must not copy. We reproduce the Objaverse-XL prompt verbatim below as the most general of the three, since it must cover humanoids, quadrupeds, avians, marine creatures, insectoids, serpentines, and articulated rigid objects within a single instruction. To ensure caption fidelity, we additionally perform a manual quality-control pass in which annotators inspect each motion sequence side-by-side with its generated caption, and revise or discard any pair in which the caption misidentifies the dominant action, attributes motion to the wrong body part, or disagrees with the underlying clip; this human-in-the-loop check guarantees that both the motion data and the textual supervision used to train \themodel{} meet a consistent quality bar.
Clips with and without root translation are both retained for training; the captions make the distinction explicit, labeling stationary motions with an ``in place'' qualifier so that the prompt disambiguates in-place articulation from root-translating locomotion.

\begin{lstlisting}
# Role
You caption clips from the Objaverse 3D animation dataset — humanoids (majority), quadrupeds, avians, marine creatures, insectoids, serpentines, and articulated rigid objects.

# Input
Four synchronized cameras 90 degrees apart at fixed elevation, labeled only by their relative azimuth about the vertical axis (0°, 90°, 180°, 270°; 0° and 180° are opposite cameras, as are 90° and 270°). The labels carry NO information about the asset's canonical heading — the subject's world orientation is arbitrary, so any camera may be seeing the front, back, side, or an oblique angle. Determine facing from the body itself (protocol step 1). The cameras are STATIC: when the subject shifts across the frame or grows/shrinks, that is the subject translating, never camera motion. A view looking straight along an elongated body may show only a compact silhouette — rely on the other views. The frames of each view are uniformly sampled across the clip in chronological order, and the views are synchronized: frame k of every view shows the same moment. Pose can change noticeably between consecutive sampled frames.

# Task
ONE short sentence describing the dominant motion in OBJECT-RELATIVE terms. Subject is 'An object', body parts use natural anatomy words (arm, wing, tail, ...). When topology is ambiguous, default to humanoid vocabulary — most assets in this dataset are humanoid.

# Observation protocol (silent — write only the caption)
1. ESTABLISH HEADING. Cues, in priority order: (a) head/face/eye/snout direction, (b) spine direction shoulders→hips for quadrupeds, (c) beak/head for avians, (d) head vs. tail end for serpentines, (e) principal translation axis for faceless rigid assets.
2. CHIRALITY. From the front, the object's left side is on the viewer's right (mirror). Apply consistently — never label a limb by which side of the camera frame it sits on.
3. SCAN every frame across all four views; do not infer from first/last alone. Note which body parts change pose and how.
4. CLASSIFY: translation (moves through space), rotation in place, articulation only (limbs/wings/tail move, body fixed), or held pose. Direction is read relative to the heading from step 1. Say 'in place' only for rotation without translation, or for locomotion-style movement without translation ('walks in place'); never as a default. Compare the subject's facing at the START and END of the clip: if it differs, a turn happened and belongs in the caption ('turns around and walks away').
5. PICK the most specific verb that fits the kinematics.

# Body-part vocabulary (descriptions of shape, not category labels)
- Humanoid/bipedal:  arm, hand, leg, foot, torso, head, hip, shoulder
- Quadruped:         front leg, hind leg, head, torso, tail
- Winged/flying:     wing, head, torso, tail, leg (if visible)
- Serpentine:        head, body, tail
- Aquatic:           fin, tail, head, body
- Insectoid:         leg, body, head
- Articulated rigid: part, segment, base, top, arm (mechanical)
If shape fits no row: upper/lower part, left/right side, front, back.

# Rules
- Format: 'An object [action].' (one sentence, <12 words)
- ONE dominant action — or a short two-phase sequence ('X and then Y') when the clip clearly has two stages.
- A cyclic motion (walk cycle, idle sway) is described ONCE as a continuous action, never per repetition; reserve the two-phase form for genuinely distinct stages.
- A concurrent pose or secondary movement may be attached with 'while ...' / 'with ...' ('walks forward with arms swinging', 'rotates its base while extending an arm').
- Mention direction (forward/backward/left/right/up/down/in place) only if clearly observed; omit rather than guess.
- All directions and chirality are object-relative (step 1), never camera-relative. If heading is uncertain, omit chirality.
- Mention a body part only when essential to disambiguate.
- No category names ('person', 'dog', 'dragon', 'car', 'robot', ...); anatomy words (arm, wing, tail) are allowed.
- No adverbs, no appearance/color/material/texture, no scene/lighting.
- Only say 'stands still' if the pose is truly unchanged across ALL frames; subtle sway, breathing, or limb shifts still count as motion.

# Style examples (format only — do NOT copy unless the motion matches)
- An object walks forward with arms swinging.
- An object kicks with the right leg while pivoting.
- An object flaps its wings and rises.
- An object slithers forward.
- An object rotates its upper segment in place.
- An object crouches down and springs upward.

Respond with ONLY the caption sentence — plain text, no quotation marks.
\end{lstlisting}

\subsection{Balanced Sampling and Augmentation}
\label{app:augmentation}
\thedata{} is heavily long-tailed: humanoids dominate while rare species are underrepresented. We mitigate this with a re-weighted sampler that assigns each instance of skeletal type $i$ (with $n_i$ samples) the weight $w_i = n_i^{-\alpha}$. Setting $\alpha = 0.5$ yields \emph{square-root sampling}, which trades off uniform sampling ($\alpha = 0$, which underexposes rare structures) against full balancing ($\alpha = 1$, which overfits tiny categories).

On top of sampling, we apply four kinematics-preserving augmentations on the fly during training, each with a fixed probability.
\emph{Joint removal} prunes a random subset of leaf joints, preferring short bones.
\emph{Joint addition} inserts a kinematically neutral joint between an existing joint and its parent, splitting the bone so that forward kinematics is preserved.
\emph{Skeleton pooling}~\cite{pool} collapses degree-2 joints by composing each single-child joint's local transform into its child, compressing linear chains without changing articulated motion.
\emph{Bone-length perturbation} scales each non-root bone offset by a factor near unity, varying limb proportions without modifying rotations.

After augmentation, motion features are recomputed through forward kinematics and checked for self-consistency; clips that fail the check are skipped. Because augmentation changes the kinematic tree, the topology descriptors of \cref{sec:representation} (Laplacian eigenvectors, graph-distance and relation matrices, and joint depths) are recomputed as well, so that the spectral coordinates and graph biases always reflect the augmented skeleton.

\section{Implementation Details}
\label{app:implementation}

\subsection{Architecture}
The motion diffusion transformer comprises $N=8$ Skeletal-Temporal Transformer blocks with hidden dimension $d=512$, $n_{\mathrm{head}}=8$ attention heads (per-head dimension $d_h=64$), and a SwiGLU~\cite{swiglu} feed-forward network of intermediate dimension $4d=2048$. We use RMS normalization~\cite{rms} and query-key normalization~\cite{qknorm,vit22b} throughout. The graph distance and relation-type embeddings have dimension $d_e=128$, and Spec-RoPE uses the $m=8$ leading non-trivial Laplacian eigenvectors with the SignNet~\cite{signnet} backend by default. The pooled topological condition $c_{\mathrm{topo}}$ is obtained by cross-attention pooling with $n_q=4$ learnable queries. Per-joint name embeddings are enabled by default; they are precomputed once on the unique joint-name vocabulary and cached. The conditioning signal, classifier-free guidance, and the AdaLN-Zero injection scheme are described in \cref{app:conditioning}.

\subsection{Conditioning}
\label{app:conditioning}
TADiT is conditioned on the diffusion timestep, the input text prompt, and the skeleton topology. The flow time $\tau$ is mapped to a timestep embedding through a sinusoidal positional encoding followed by an MLP,
\begin{equation}
c_\tau = \mathrm{MLP}(\mathrm{PE}(\tau)).
\end{equation}
The text prompt is encoded into a latent text condition $c_{\mathrm{text}}$ by a frozen FLAN-T5-Base~\cite{flan-t5} encoder, whose output token sequence is linearly projected to dimension $d=512$. To enable classifier-free guidance~\cite{cfg}, $c_{\mathrm{text}}$ is independently dropped to a learnable null embedding with probability $p_{\mathrm{cf}}$ during training. The topology-specific term $c_{\mathrm{topo}}$ is defined in the main paper. The three condition vectors are fused into a single global signal $c = c_\tau + c_{\mathrm{text}} + c_{\mathrm{topo}}$, which is injected into every transformer sublayer (joint attention, temporal attention, MLP) and the output head through adaptive layer normalization with zero-initialized gating (AdaLN-Zero)~\cite{dit}. For a sublayer $f$ acting on a hidden state $h$, AdaLN-Zero produces
\begin{equation}
h' = h + \alpha(c) \odot f\!\bigl((1 + \gamma(c)) \odot \mathrm{LN}(h) + \beta(c)\bigr),
\label{eq:adaln-zero}
\end{equation}
where $(\alpha, \beta, \gamma) = W_c\,c$ are zero-initialized linear projections of the conditioning vector, so $\alpha\!=\!0$ at initialization and each block starts from the identity transform.

\subsection{Output Head}
After the final transformer block, the latent tensor takes the shape $Z^{(N)} \in \mathbb{R}^{(T+1) \times J \times d}$. The output head maps latent tokens back to the motion feature space. Because root and non-root joints follow distinct feature conventions, we decode them with separate two-layer MLPs, both AdaLN-Zero modulated by $c$. Prior to decoding, the root token at each frame attends to the non-root joint tokens of the same frame through a zero-initialized cross-attention block, aggregating whole-body information into the root channel as needed. Letting $\Pi(\cdot)$ denote the final decoder, the predicted velocity field is $\hat{V} = \Pi(Z^{(N)}, c) \in \mathbb{R}^{T\times J\times D}$. The prepended topology slot is discarded after decoding, and the remaining output is reshaped to the original motion layout.

\subsection{Training Objective}
\label{app:training-objective}
The masked flow-matching MSE loss is
\begin{equation}
\mathcal{L}_{\mathrm{mse}}
=
\mathbb{E}_{x_1,\, x_0,\, \tau}
\!\left[
\frac{\bigl\| \Omega \odot \bigl(v_\theta(x_\tau, \tau, c, \mathcal{S}) - (x_1 - x_0)\bigr)\bigr\|_2^2}{\|\Omega\|_1}
\right],
\end{equation}
with $\tau \sim \mathcal{U}(0,1)$ and $\Omega \in \{0,1\}^{T\times J\times D}$ a binary validity mask that zeros out padded joint slots in heterogeneous-skeleton batches.

The geodesic loss on $SO(3)$ is applied to the one-step denoised rotations rather than to the velocity itself, so that supervision is performed in the clean motion space. From the predicted velocity, the clean motion estimate is $\hat{x}_1 = x_\tau + (1-\tau)\, v_\theta(x_\tau, \tau, c, \mathcal{S})$. Let $\hat{R}_j^t \in SO(3)$ denote the rotation matrix recovered from the 6D rotation channels of $\hat{x}_1$ at frame $t$ and joint $j$, and $R_j^t \in SO(3)$ the corresponding ground-truth rotation. Then
\begin{equation}
\begin{aligned}
\mathcal{L}_{\mathrm{geo}}
&=
\mathbb{E}_{x_1,\, x_0,\, \tau}
\!\left[
\frac{\sum_{j,t} \omega_{j,t}\, d_{\mathrm{geo}}\!\bigl(\hat{R}_j^t,\, R_j^t\bigr)}{\sum_{j,t} \omega_{j,t}}
\right],
\\
d_{\mathrm{geo}}(R_a, R_b)
&=
\arccos\!\left(\frac{\mathrm{tr}\!\bigl(R_a^{\!\top} R_b\bigr) - 1}{2}\right),
\end{aligned}
\end{equation}
where $\omega_{j,t} \in \{0,1\}$ is the per-joint validity mask induced by $\Omega$ and the $\arccos$ argument is clamped to $[-1+\epsilon,\, 1-\epsilon]$ for numerical stability.

The velocity smoothness regularizer applies to the temporal acceleration of the denoised velocity channels. Letting $\hat{v}_j^t \in \mathbb{R}^3$ denote the velocity block of $\hat{x}_1$ at joint $j$ and frame $t$,
\begin{equation}
\mathcal{L}_{\mathrm{smooth}}
=
\mathbb{E}_{x_1,\, x_0,\, \tau}
\!\left[
\frac{1}{3\sum_{j,t} \omega_{j,t}^{\Delta}}
\sum_{j,t} \omega_{j,t}^{\Delta}\,
\bigl\|\hat{v}_j^{t+1} - \hat{v}_j^{t}\bigr\|_2^{2}
\right],
\end{equation}
where $\omega_{j,t}^{\Delta} = \omega_{j,t}\, \omega_{j,t+1}$ restricts the finite difference to pairs of consecutive valid frames within a clip.

\subsection{Training}
We train \themodel{} with the flow-matching objective in \cref{app:training-objective}, weighting the geodesic regularizer at $\lambda_{\mathrm{geo}}=0.5$ and the smoothness regularizer at $\lambda_{\mathrm{smooth}}=0.1$. Optimization uses AdamW~\cite{adamw} with a learning rate of $1\times 10^{-4}$, $(\beta_1,\beta_2)=(0.9,0.99)$, weight decay $1\times 10^{-5}$, and gradient clipping at norm $1.0$. The learning rate follows a cosine schedule with linear warmup over the first $3\%$ of training and a floor of $5\%$ of the peak rate. We maintain an exponential moving average of model parameters with a step-dependent decay
\begin{equation}
  \eta_n = \min\!\left(0.9999,\, \frac{1+n}{10+n}\right),
  \label{eq:ema-decay}
\end{equation}
where $n$ is the optimization step, so that the EMA tracks fast updates early and saturates to $0.9999$ later in training. Training is distributed across $8$ NVIDIA H100 GPUs using bfloat16 mixed precision, with a per-GPU batch size of $32$ for a global batch size of $256$. The classifier-free guidance dropout probability is $p_{\mathrm{cf}}=0.1$ on the text condition. We train for $100$k optimization steps in total, which corresponds to approximately one day of wall-clock time on the configuration above.

\subsection{Inference}
At inference time we evaluate the EMA copy of the model. Given a rigged 3D asset and a text prompt, we first canonicalize the skeleton, compute its topology descriptors (rest-pose joint positions, graph distance and relation matrices, joint depths, Laplacian eigenvectors, and joint-name embeddings), and encode the text prompt with the frozen FLAN-T5 encoder; these conditioning tensors are computed once per asset/prompt pair and reused across samples. We then draw an initial noise tensor $x_0\sim\mathcal{N}(0,I)$ in the padded motion shape and integrate the learned velocity field $v_\theta(x_\tau,\tau,c,\mathcal{S})$ from $\tau=0$ to $\tau=1$ with a fixed-step Euler ODE solver using $50$ steps, which we found to be sufficient for visually clean trajectories. Classifier-free guidance is applied at every solver step: we run one forward pass with the text condition active and one with the null embedding, and combine them as
\begin{equation}
\hat{v}_\theta = v_\theta^{\mathrm{uncond}} + s\,(v_\theta^{\mathrm{cond}} - v_\theta^{\mathrm{uncond}}),
\end{equation}
with guidance scale $s=3.0$ by default; the unconditional and conditional branches are batched into a single forward call to avoid wall-clock overhead. After the final solver step we discard the prepended skeleton tokens, slice off the padded joint slots using the joint-validity mask, and de-normalize the predicted features with the dataset-level statistics computed during preprocessing. Joint rotations are converted from the continuous 6D representation to $SO(3)$ matrices, the root trajectory is reconstructed by integrating the yaw-canonical root velocities and re-applying the initial facing direction, and the remaining global joint poses are obtained through forward kinematics. The resulting skeleton-space motion is finally driven onto the input mesh via linear blend skinning to produce the rendered animation. Sampling a $60$-frame clip on a single NVIDIA H100 GPU takes roughly $1.2$ seconds end-to-end, including text encoding and forward-kinematic recovery.

\subsection{Animation Module}
From $\hat{x}_{1}$, we recover the global root trajectory and per-joint rotations and apply forward kinematics with the rest-pose offsets to obtain per-joint rigid transformations in $SE(3)$, which drive the mesh via linear blend skinning~\cite{lbs} or dual-quaternion blending~\cite{dqb}.

\section{Theoretical Analysis of Spec-RoPE}
\label{app:spectral-rope-theory}

This section discusses several useful properties of Spec-RoPE, the spectral rotary positional encoding employed in our topology-aware motion transformer. Unlike standard RoPE, which is defined on canonical 1D or 2D coordinates, our setting operates on the kinematic tree induced by a skeleton. We therefore construct rotary coordinates from the graph spectrum of that skeleton.

\subsection{Setup}
Let $\mathcal{T}_{\mathcal{S}}=(\mathcal{V},\mathcal{E})$ denote the kinematic tree of a skeleton with $|\mathcal{V}|=J$ joints. Let $A\in\{0,1\}^{J\times J}$ be its adjacency matrix. The combinatorial graph Laplacian is
\begin{equation}
L_{\mathcal{S}} = \mathrm{diag}(A\mathbf{1}) - A,
\end{equation}
which is symmetric and positive semidefinite, and admits the eigendecomposition
\begin{equation}
L_{\mathcal{S}} = U \Lambda U^\top,
\qquad
U = [u_0, u_1, \ldots, u_{J-1}],
\end{equation}
with $\Lambda=\mathrm{diag}(\lambda_0,\ldots,\lambda_{J-1})$ and $0=\lambda_0 \le \lambda_1 \le \cdots \le \lambda_{J-1}$. Since the kinematic tree is connected, $\lambda_0$ is simple and $u_0$ is constant; we therefore discard $u_0$ and form the spectral coordinate of joint $i$ from the leading $m$ non-trivial eigenvectors,
\begin{equation}
\mathbf{s}_i = [u_1(i),u_2(i),\ldots,u_m(i)] \in \mathbb{R}^m.
\end{equation}
Given a per-head query or key vector $z_i\in\mathbb{R}^{d_h}$, Spec-RoPE takes the block-diagonal form
\begin{equation}
\begin{aligned}
\mathrm{RoPE}(\mathbf{s}_i)\, z_i
&=
\bigoplus_{n=1}^{d_h/2}
\rho(\theta_n(\mathbf{s}_i))\,[z_i]_{2n-2:2n-1},
\\
\rho(\theta)
&=
\begin{bmatrix}
\cos\theta & \sin\theta\\
-\sin\theta & \cos\theta
\end{bmatrix},
\end{aligned}
\end{equation}
where $\theta_n:\mathbb{R}^m\to\mathbb{R}$ is a learned per-channel angle map. The analysis below studies the analytically tractable \emph{linear} parameterization
\begin{equation}
\theta_n(\mathbf{s}) = \omega_n^\top \mathbf{s}, \qquad \omega_n\in\mathbb{R}^m,
\label{eq:linear-angles}
\end{equation}
which admits the cleanest algebraic structure. The SignNet backend, used as our default, replaces \eqref{eq:linear-angles} with
the sign-symmetric construction of the main paper: each spectral coordinate is passed through a shared MLP at both signs and the symmetrized features are concatenated and mapped to the angles,
\begin{equation}
\theta_n(\mathbf{s})
=
\psi_n\!\left(\mathrm{Concat}\bigl(\{\phi(s_k)+\phi(-s_k)\}_{k=1}^{m}\bigr)\right),
\end{equation}
which is invariant to independent sign flips of the individual eigenvectors;
this trades the strict relative-coordinate property below for invariance to the spectral sign ambiguity, while preserving the qualitative behavior.

\subsection{Proposition 1 (Translation Invariance in Spectral Coordinates)}
\emph{Under the linear parameterization \eqref{eq:linear-angles}, the inner product between rotary-encoded queries and keys depends on the spectral coordinates only through their difference:}
\begin{equation}
\begin{split}
\mathrm{RoPE}(\mathbf{s}_i)^\top \mathrm{RoPE}(\mathbf{s}_j)
&=
\mathrm{RoPE}(\mathbf{s}_j-\mathbf{s}_i),
\\
\langle \mathrm{RoPE}(\mathbf{s}_i)\,q,\, \mathrm{RoPE}(\mathbf{s}_j)\,k \rangle
&=
q^\top \mathrm{RoPE}(\mathbf{s}_j-\mathbf{s}_i)\,k.
\end{split}
\end{equation}
\textbf{Proof.}
Each $2\times 2$ block satisfies $\rho(\alpha)^\top\rho(\beta)=\rho(\beta-\alpha)$ since $\rho$ is a one-parameter subgroup with $\rho(\alpha)^\top=\rho(-\alpha)$. With linear angles, $\omega_n^\top \mathbf{s}_j-\omega_n^\top \mathbf{s}_i=\omega_n^\top(\mathbf{s}_j-\mathbf{s}_i)$, so each block reduces to $\rho(\omega_n^\top(\mathbf{s}_j-\mathbf{s}_i))$. The block-diagonal structure of $\mathrm{RoPE}$ then yields the matrix identity, and the inner-product form follows immediately. $\square$

Consequently the rotary encoding injects \emph{relative} structural information from the kinematic tree into attention, mirroring the central design principle of standard RoPE on regular sequences.

\subsection{Proposition 2 (Permutation Equivariance)}
A spectral graph encoding should depend on the kinematic tree itself, not on any particular enumeration of its joints. Our construction satisfies this property up to the standard ambiguities intrinsic to eigendecompositions.

\emph{Let $P\in\mathbb{R}^{J\times J}$ be the permutation matrix induced by a reordering of the joints. Then the reordered Laplacian satisfies $L'_{\mathcal{S}}=P\, L_{\mathcal{S}}\, P^\top$. Its eigenvalues coincide with those of $L_{\mathcal{S}}$, and its eigenvectors transform as $u'_k=\pm P u_k$ when $\lambda_k$ is simple, and as $U'_\Lambda = P\, U_\Lambda\, Q$ for some $Q\in O(\dim \mathcal{E}_\Lambda)$ within any degenerate eigenspace $\mathcal{E}_\Lambda$. Consequently, Spec-RoPE is equivariant to joint permutation modulo this standard sign-and-basis ambiguity.}

\textbf{Proof.}
Permuting the joint set conjugates both the adjacency and degree matrices by $P$, hence $L'_{\mathcal{S}}=P L_{\mathcal{S}} P^\top$. Conjugation preserves the spectrum, so the eigenvalues are invariant. If $L_{\mathcal{S}} u_k=\lambda_k u_k$ then $L'_{\mathcal{S}}(P u_k)=P L_{\mathcal{S}} P^\top P u_k=\lambda_k (P u_k)$, identifying $P u_k$ as an eigenvector of $L'_{\mathcal{S}}$. For a simple eigenvalue, eigenvectors are determined up to sign, giving $u'_k=\pm P u_k$. For a degenerate eigenvalue, any orthonormal basis of the eigenspace is admissible, leaving the residual orthogonal freedom $Q\in O(\dim\mathcal{E}_\Lambda)$. The spectral coordinates $\mathbf{s}_i$ inherit the same equivariance up to these intrinsic ambiguities, and the SignNet backend further removes the sign ambiguity by construction: writing $\pi$ for the relabeling, simple $\lambda_1,\dots,\lambda_m$ give $\mathbf{s}'_{\pi(i)}=[\pm u_1(i),\dots,\pm u_m(i)]$, and since $\phi(s)+\phi(-s)$ is even in each coordinate, the SignNet angles satisfy $\theta_n(\mathbf{s}'_{\pi(i)})=\theta_n(\mathbf{s}_i)$, so the rotary matrices permute with the joints and the attention logits are consistent under relabeling. $\square$

\paragraph{Repeated and near-repeated eigenvalues.}
SignNet resolves the independent sign ambiguity of each eigenvector, but it is not invariant to an arbitrary orthogonal change of basis within a repeated eigenspace. Near-repeated eigenvalues can likewise yield numerically unstable eigenvectors: small perturbations may rotate their basis substantially even when the underlying subspace changes little. Spec-RoPE alone therefore does not provide complete basis invariance in these cases. The truncation at $m$ inherits the same caveat: when $\lambda_m=\lambda_{m+1}$, the cut falls inside a repeated eigenspace, so even the subspace spanned by the leading $m$ non-trivial eigenvectors is ambiguous. In the full architecture, however, spectral coordinates are only one of several complementary structural cues. Graph distance and relation-type biases are independent of the eigenspace basis, while rest-pose geometry, joint depth, and joint-name embeddings provide additional geometric, hierarchical, and semantic information. The model can therefore retain structural identifiability when individual spectral axes are ambiguous rather than relying exclusively on their orientation. A fully basis-invariant encoding of degenerate eigenspaces remains an interesting direction for future work.

\subsection{Proposition 3 (Connection to Standard RoPE)}
Spec-RoPE generalizes ordinary RoPE from regular grids to arbitrary skeletal graphs.

\emph{When the underlying graph is a path graph on $J$ vertices, the first non-trivial Laplacian eigenvector $u_1$ is a strictly monotone function of the token index, and, restricted to this leading spectral coordinate, Spec-RoPE recovers standard 1D RoPE up to a monotone reparameterization of the position coordinate. More generally, on Cartesian products of path graphs the axis-aligned first-harmonic eigenvectors vary along one axis each and recover the multi-axis RoPE used in image and video transformers, up to per-axis reparameterization.}

\textbf{Proof.}
For a path graph $P_J$ the Laplacian eigenpairs admit the closed form $u_k(j)\propto\cos\!\big((j+\tfrac{1}{2})\pi k/J\big)$ with $\lambda_k=2-2\cos(\pi k/J)$ for $k=0,\ldots,J-1$~\cite{graph-transformer,graphgps}. The first non-trivial eigenvector $u_1$ is therefore a strictly monotone function of the index $j$, so an angle map supported on the leading coordinate, $\theta_n=\omega_{n,1}\,u_1(j)$, is a monotone reparameterization of the standard rotary angle $\omega_{n,1}\,j$; the higher coordinates $u_2,\dots,u_m$ oscillate in $j$, so the reduction concerns this leading component. For a Cartesian product $P_{J_1}\times\cdots\times P_{J_d}$, the Laplacian eigenvectors factorize as $u_{k_1,\ldots,k_d}(j_1,\ldots,j_d)=\prod_a u_{k_a}(j_a)$, so the axis-aligned first harmonics vary along one axis each and play the role of axis-wise positional coordinates; which eigenvectors are \emph{leading} depends on the side lengths---for strongly skewed products, several higher harmonics of the longest axis precede the first harmonic of a shorter axis. In both cases Spec-RoPE thus reduces to existing RoPE constructions under a smooth, monotone change of coordinates rather than introducing a fundamentally different mechanism. $\square$

\subsection{Interpretation via Effective Resistance}
A useful way to understand the structural bias induced by Spec-RoPE is through the effective resistance of the kinematic tree. Let $L_{\mathcal{S}}^\dagger$ denote the Moore--Penrose pseudoinverse of the Laplacian. The effective resistance between joints $i$ and $j$ is
\begin{equation}
R_{\mathrm{eff}}(i,j)
=
(L_{\mathcal{S}}^\dagger)_{ii}
+
(L_{\mathcal{S}}^\dagger)_{jj}
-
2(L_{\mathcal{S}}^\dagger)_{ij},
\end{equation}
and the spectral expansion $L_{\mathcal{S}}^\dagger=\sum_{k=1}^{J-1}\lambda_k^{-1} u_k u_k^\top$ yields
\begin{equation}
R_{\mathrm{eff}}(i,j)
=
\sum_{k=1}^{J-1}
\frac{(u_k(i)-u_k(j))^2}{\lambda_k}.
\end{equation}
Defining the scaled spectral coordinate
\begin{equation}
\tilde{\mathbf{s}}_i
=
\left[
\frac{u_1(i)}{\sqrt{\lambda_1}},\,
\frac{u_2(i)}{\sqrt{\lambda_2}},\,
\ldots,\,
\frac{u_{J-1}(i)}{\sqrt{\lambda_{J-1}}}
\right],
\end{equation}
this becomes the squared Euclidean distance in the scaled spectral embedding,
\begin{equation}
R_{\mathrm{eff}}(i,j)=\|\tilde{\mathbf{s}}_i-\tilde{\mathbf{s}}_j\|_2^2.
\end{equation}
In other words, differences in scaled spectral coordinates encode a graph-geometric distance. For a tree with unit edge weights---which our kinematic skeletons are---effective resistance coincides with shortest-path distance, so joints farther apart along the articulated hierarchy are also farther apart in scaled spectral space.

\subsection{Small-Variance Interpretation}
The effective-resistance identity offers a transparent picture of how rotary attention behaves in scaled spectral space. Suppose we use scaled coordinates $\tilde{\mathbf{s}}_i$ in place of $\mathbf{s}_i$ and average the rotary kernel over isotropic Gaussian frequencies $\omega\sim\mathcal{N}(0,\sigma^2 I_{J-1})$. Standard properties of the characteristic function of a Gaussian give the closed form
\begin{equation}
\begin{split}
\mathbb{E}_\omega\!\left[\cos\!\big(\omega^\top(\tilde{\mathbf{s}}_i-\tilde{\mathbf{s}}_j)\big)\right]
&=
\exp\!\left(-\tfrac{\sigma^2}{2}\,\|\tilde{\mathbf{s}}_i-\tilde{\mathbf{s}}_j\|_2^2\right)
\\
&=
\exp\!\left(-\tfrac{\sigma^2}{2}\,R_{\mathrm{eff}}(i,j)\right),
\end{split}
\end{equation}
so the expected rotary kernel decays exponentially in the effective resistance, and---to leading order in $\sigma^2$---its first-order correction is proportional to $R_{\mathrm{eff}}(i,j)$. This motivates viewing Spec-RoPE as a soft structural prior that suppresses attention between joints that are far apart on the kinematic tree.

In our model this picture should be read as an interpretive guide rather than a literal theorem about the deployed architecture, for three reasons: (i) we use a truncated set of $m$ leading spectral coordinates rather than the full spectrum; (ii) the angles are produced by a learned deterministic map---linear in the analytically tractable case, sign-symmetric and nonlinear in our default SignNet construction---rather than by averaging over random Gaussian frequencies; and (iii) attention additionally incorporates an explicit graph bias through relation- and distance-type embeddings. The closed-form expectation therefore does not transfer verbatim. Empirically, however, the qualitative effect persists: rotations driven by spectral coordinates encourage attention to depend on relative graph geometry rather than on an arbitrary joint ordering.

\subsection{Summary}
Taken together, Propositions~1--3 justify our use of Spec-RoPE in a topology-aware motion model. The encoding is well-defined on arbitrary kinematic trees, is insensitive to joint permutation up to intrinsic spectral ambiguities (with SignNet removing eigenvector sign ambiguity), reduces to standard RoPE on regular structures, and admits a natural distance-based interpretation on trees through effective resistance. These properties make it particularly suitable for our setting, where a single model must generalize across heterogeneous skeleton topologies while preserving meaningful structural relationships between joints.

\section{Experimental Protocols}
\label{app:experiments}

This section documents the evaluation protocols behind the experiments in the main paper. \cref{app:evaluation-metrics} distinguishes the metrics used for skeletal motion and rendered mesh animation, and \cref{app:heldout} lists the exact held-out skeletons, per-sequence prompts, and random seed. \cref{app:ood} describes our out-of-distribution evaluation, \cref{app:baselines} discusses the compared methods and the adaptations required for a fair comparison, and \cref{app:representation-tradeoff} summarizes the complementary strengths of skeleton-based and skeleton-free animation. Finally, \cref{app:user-study} details the protocol of our anonymous perceptual study.

\subsection{Evaluation Metrics}
\label{app:evaluation-metrics}
We evaluate the two comparison settings in their native output domains. For skeleton-based motion generation, motion quality is measured directly in skeleton space using FID over kinematic features, together with diversity over generated joint trajectories. VBench is used only for the mesh-animation comparison: following the protocol of the skeleton-free baselines~\cite{animateanymesh,animax}, we render all methods as fixed-view videos and evaluate their perceptual quality in video space. This separation avoids using a video metric as a proxy for skeletal motion quality while retaining a common output representation for methods that do not produce skeletons.

\subsection{Held-Out Evaluation Set}
\label{app:heldout}

The motion-generation comparison in the main paper holds out seven Truebones skeleton types that span six morphology categories: Raptor (bipedal), Leapord and Raindeer (quadrupedal), Parrot2 (avian), Jaws (marine), Crab (insectoid), and KingCobra (serpentine). No motion of these skeletons is seen during training. Together they contribute 58 evaluation sequences; \cref{tab:heldout-prompts} lists, for each skeleton type, the exact text prompts used to condition generation. Skeleton names keep the original Truebones spelling (e.g., \texttt{Leapord}, \texttt{Raindeer}), while prompts refer to each character by its species name. All quantitative results are generated with random seed 10.

\begin{table*}[t]
  \centering
  \caption{\textbf{The held-out evaluation set.} The 58 evaluation prompts of the seven held-out Truebones skeletons (Crab 10, Jaws 8, KingCobra 8, Parrot2 2, Leapord 12, Raindeer 8, Raptor 10), grouped by skeleton type; each prompt is the exact text used to condition generation.}
  \label{tab:heldout-prompts}
  \scriptsize
  \begin{tabular}{@{}ll@{\hskip 14pt}ll@{}}
    \toprule
    \textbf{Skeleton} & \textbf{Prompt} & \textbf{Skeleton} & \textbf{Prompt} \\
    \midrule
    \texttt{Crab} & A crab attacks forward. & \texttt{Leapord} & A leopard attacks forward. \\
    & A crab lunges forward to attack. & & A leopard crouches low and bites forward. \\
    & A crab strikes forward with its claws. & & A leopard collapses and lies on its side. \\
    & A crab snaps its claws and attacks. & & A leopard falls backward and lies on its side. \\
    & A crab lowers its head and lies down. & & A leopard crouches low. \\
    & A crab collapses backward and lies down. & & A leopard growls with its head lowered. \\
    & A crab eats. & & A leopard gets up from lying down. \\
    & A crab lies on its back. & & A leopard pounces forward. \\
    & A crab strafes. & & A leopard pounces forward and swipes with its front legs. \\
    & A crab walks forward using its legs. & & A leopard runs forward. \\
    \texttt{Parrot2} & A parrot preens its feathers. & & A leopard stands still. \\
    & A parrot flaps its wings and glides forward. & & A leopard walks forward. \\
    \addlinespace
    \texttt{Raindeer} & A reindeer attacks forward and strikes with its head. & \texttt{KingCobra} & A king cobra strikes forward with tail curled. \\
    & A reindeer charges forward with its antlers lowered. & & A king cobra rears up and strikes forward. \\
    & A reindeer thrusts its head forward to attack. & & A king cobra strikes forward. \\
    & A reindeer falls backward and lies on its side. & & A king cobra circles and bites its own tail. \\
    & A reindeer runs forward. & & A king cobra raises its head. \\
    & A reindeer stands still. & & A king cobra stands still. \\
    & A reindeer walks forward. & & A king cobra raises its head and holds a still pose. \\
    & A reindeer lowers its head and snorts. & & A king cobra throws its head back and falls forward. \\
    \addlinespace
    \texttt{Jaws} & A shark bites to the left. & \texttt{Raptor} & A raptor attacks forward and strikes with its head. \\
    & A shark bites to the right. & & A raptor falls backward and lies on its side. \\
    & A shark thrashes in place with its jaws open. & & A raptor walks forward at a brisk pace. \\
    & A shark swims forward with its jaws open. & & A raptor lowers its head and lunges forward. \\
    & A shark swims forward and turns around. & & A raptor stands still. \\
    & A shark swims forward. & & A raptor holds a still standing pose. \\
    & A shark swims to the left. & & A raptor scratches with its right foreleg. \\
    & A shark swims to the right. & & A raptor roars while standing still. \\
    & & & A raptor walks forward slowly. \\
    & & & A raptor yawns with its head raised upward. \\
    \bottomrule
  \end{tabular}
\end{table*}

\subsection{Out-of-Distribution Evaluation}
\label{app:ood}
In addition to the held-out Truebones~\cite{truebones} skeletons and the mesh-animation benchmark of the main paper, we evaluate \themodel{} on a small curated set of AI-generated and automatically rigged meshes. These assets exhibit non-canonical proportions, irregular skeleton topologies, and rigging conventions that are not represented in any of our training sources, and therefore probe the model's ability to generalize beyond artist-authored rigs.

\subsection{Baselines}
\label{app:baselines}
\paragraph{AnyTop.}
For a fair comparison, we extend AnyTop~\cite{anytop} with the cross-attention text-conditioning module described in the main paper, since the published model is purely unconditional. Beyond this adaptation, three structural limitations of AnyTop are worth highlighting. \emph{(i)} It relies on skeleton-specific statistical estimation and normalization, which requires the user to supply at least one reference motion sequence for every target skeleton before generation can begin. \emph{(ii)} It is trained on a comparatively small corpus of artist-crafted animal skeletons, which constrains the topological diversity it can express. \emph{(iii)} It is fundamentally an unconditional generator and therefore lacks any native mechanism for textual control, which limits its practical utility for user-driven animation. Our cross-attention extension partly addresses (iii) for the purpose of comparison, but does not remedy (i) or (ii).

\paragraph{How to Move Your Dragon.}
Closest to our text-conditioned setting, How to Move Your Dragon~\cite{t2m4lvo} annotates the Truebones corpus with text descriptions and trains a topology-adaptive motion diffusion model on it, using rig augmentation to broaden the skeletal configurations seen during training. It therefore natively addresses limitation (iii) above, but shares limitation (ii): its training domain remains the same small, artist-crafted animal corpus, whereas \thedata{} spans humans, animals, and general articulated objects over thousands of distinct skeletons. To our knowledge, neither code nor pretrained models had been released at the time of writing, so we do not include it in our quantitative comparison.

\paragraph{AnimaX}
We do not include AnimaX~\cite{animax} in our mesh-animation comparison: the authors have not publicly released their code or pretrained models, and the technical report does not provide enough detail for a faithful reimplementation.

\subsection{Skeleton-Based vs. Skeleton-Free Animation}
\label{app:representation-tradeoff}
The two paradigms make complementary representation choices. A skeletal representation is compact and semantically meaningful: a motion is expressed as a small set of joint transformations rather than dense vertex trajectories. It decouples motion from a particular mesh geometry, allowing one generated sequence to drive different skinned assets with compatible articulation, and it is efficient to edit using established rig-based tools. It also integrates naturally with inverse kinematics, physics simulators, motion-capture pipelines, and interactive controllers. These properties make skeleton-based methods especially suitable for articulated characters and production workflows that require reusable and controllable motion. Their main cost is the need for a valid rig, and the kinematic tree constrains deformation to motions that its joints and skinning model can express.

Skeleton-free methods instead predict deformation directly in mesh, point, or implicit geometry space. By avoiding a fixed joint hierarchy, they can better accommodate objects that are not naturally articulated, highly non-rigid deformation, and fine-grained surface dynamics that would require an impractically dense rig. This flexibility comes with a substantially higher-dimensional output, weaker built-in kinematic structure, and less direct compatibility with standard rig-based editing and control. Thus, skeleton-free methods are not superseded by our formulation; they are preferable when deformation cannot be captured well by a kinematic skeleton, whereas \themodel{} targets efficient, controllable animation of rigged assets. Hybrid models that retain skeletal control while learning residual non-rigid deformation are a promising direction for combining both strengths; \cref{app:skeleton-free} contrasts the two paradigms visually.

\subsection{User Study}
\label{app:user-study}
\begin{figure}[!h]
  \centering
  \begin{subfigure}[b]{ 1.0\linewidth}
    \centering
    \includegraphics[width=\linewidth]{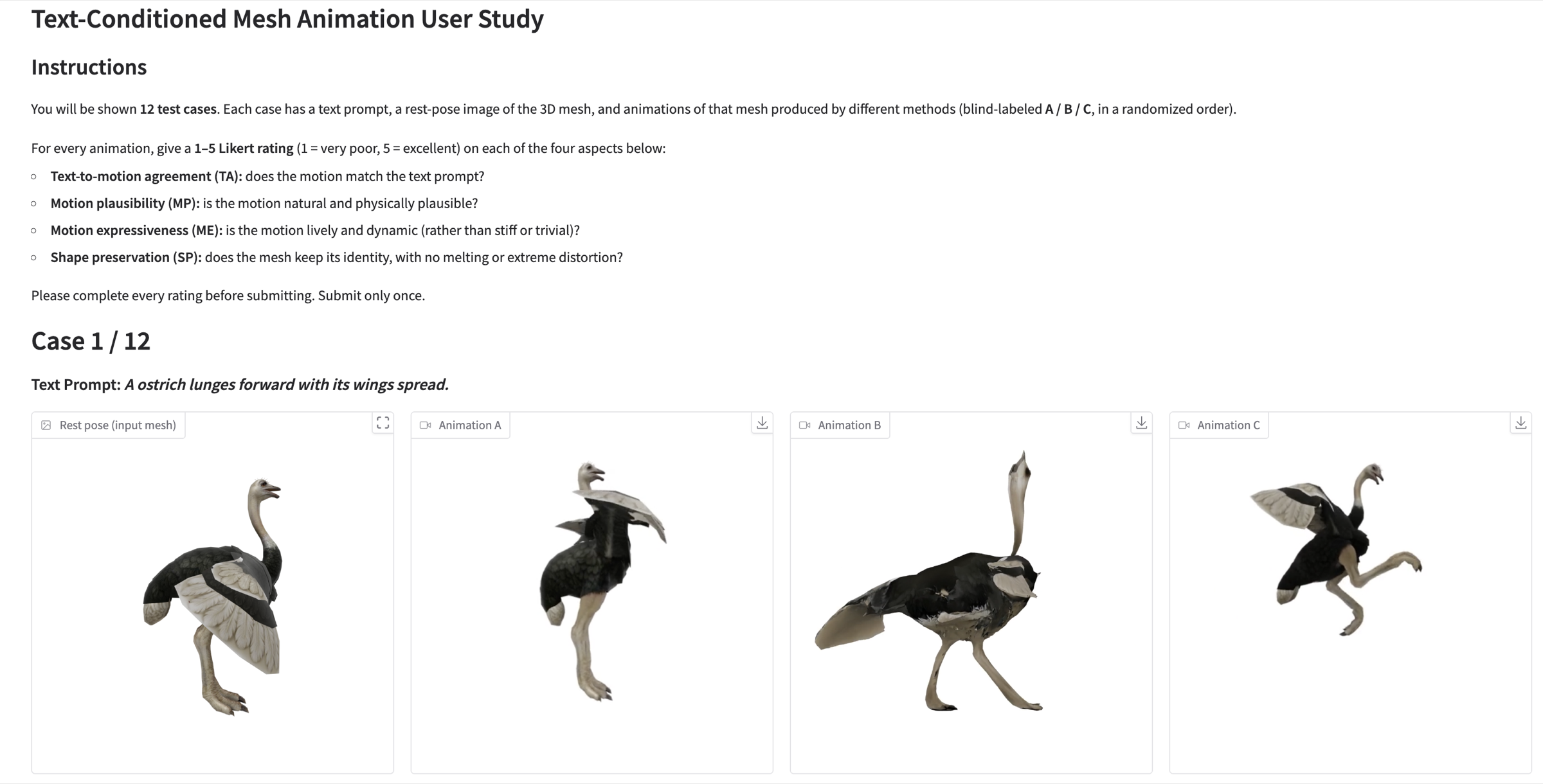}
    \Description{User study instruction interface explaining the animation comparison task.}
    \caption{Participant instruction page.}
    \label{fig:user-study-instructions}
  \end{subfigure}
  \begin{subfigure}[b]{ 1.0\linewidth}
    \centering
    \includegraphics[width=\linewidth]{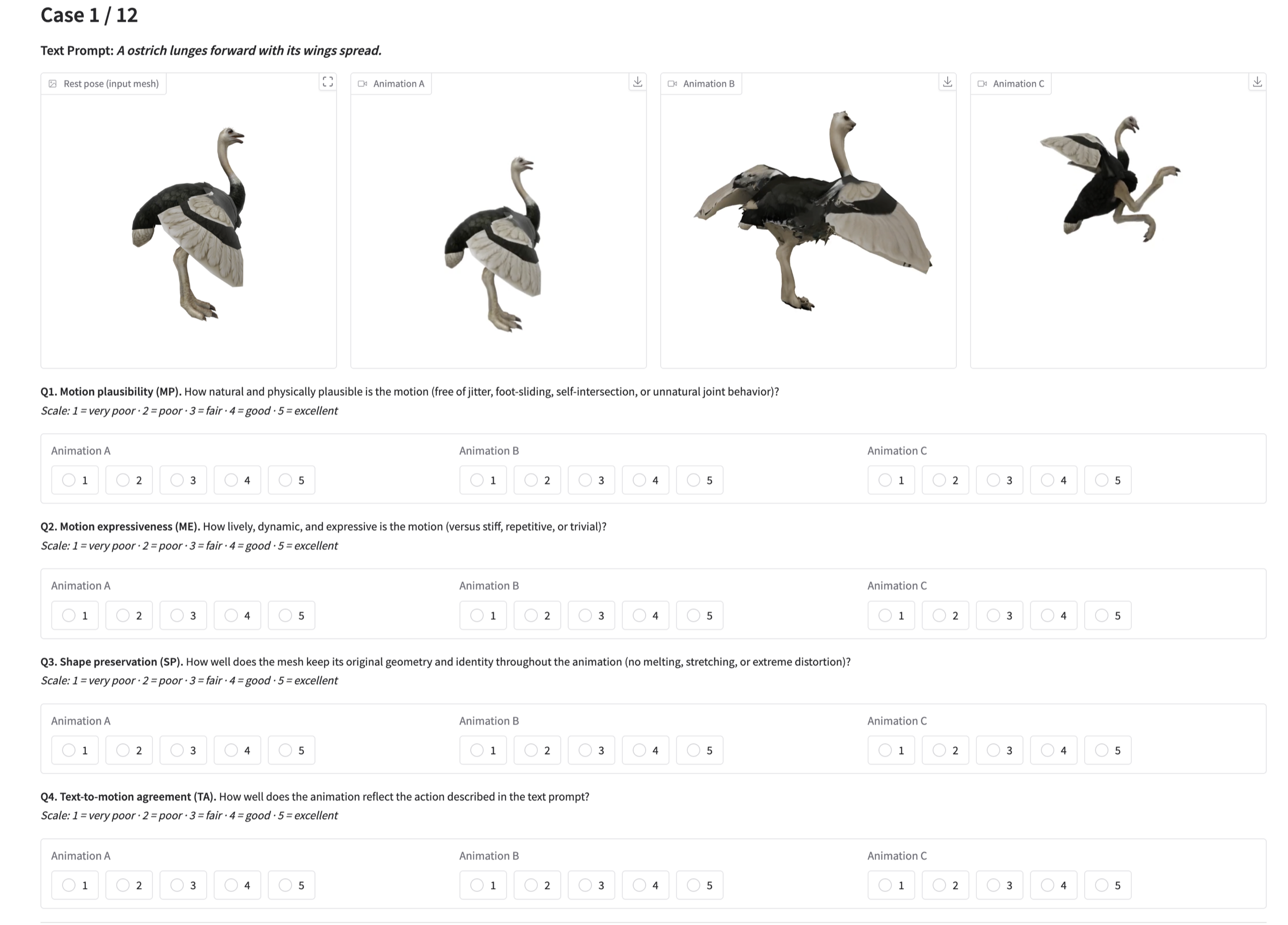}
    \Description{Representative user study trial presenting animations for comparison.}
    \caption{Representative test trial.}
    \label{fig:user-study-question}
  \end{subfigure}
  \caption{\textbf{User-study interface and protocol.} (a)~Instruction page presented at the start of each session, with institution-identifying information masked. (b)~A representative trial: three side-by-side videos generated by different methods from the same input mesh and text prompt, followed by the four evaluation questions used for subjective scoring; method--label assignments are anonymized and randomized per participant.}
  \label{fig:user-study-form}
\end{figure}

\begin{figure}[t]
  \centering
  \includegraphics[width= 1.0\linewidth]{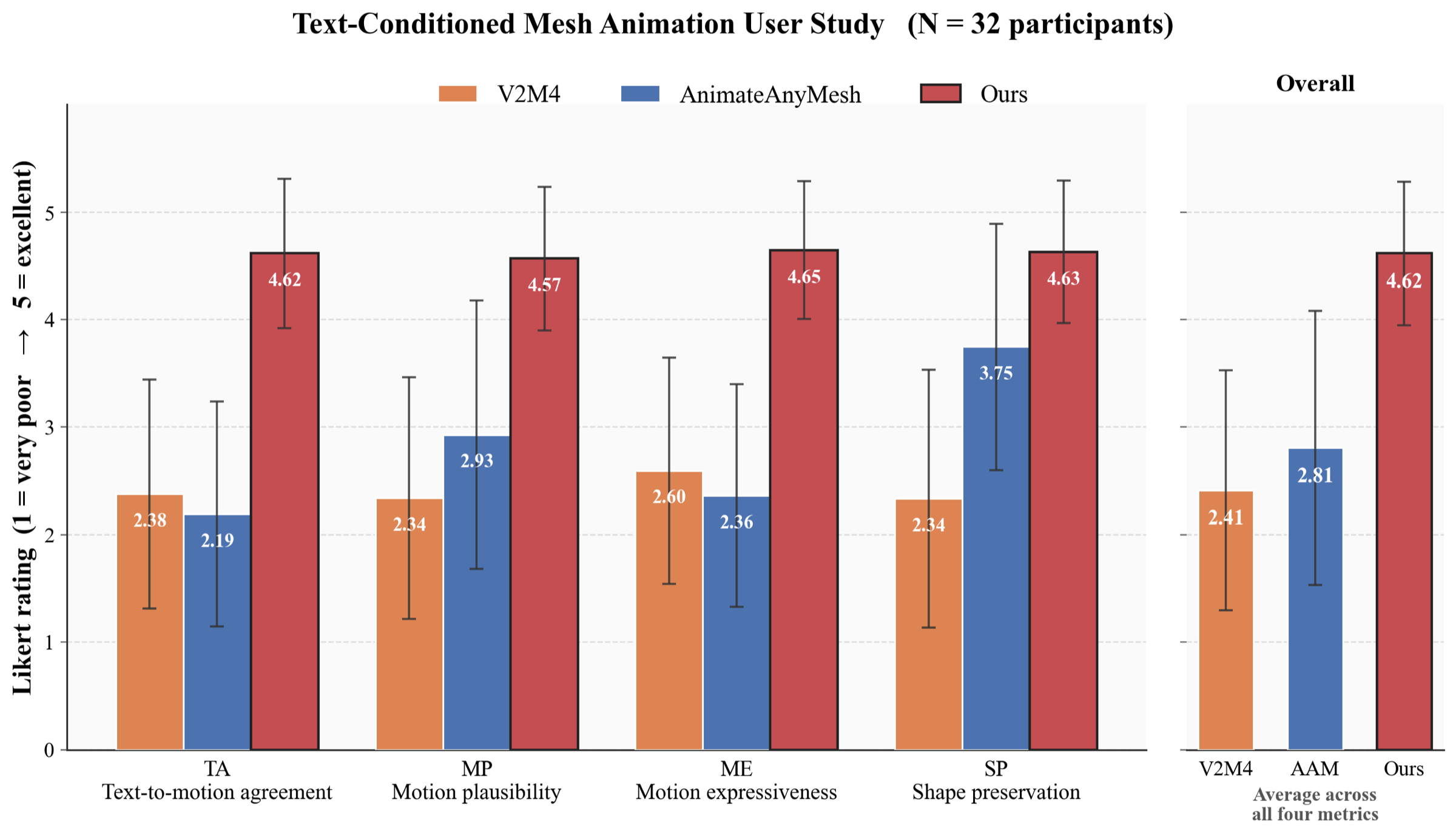}
  \Description{Grouped bar chart of mean Likert ratings for V2M4, AnimateAnyMesh, and UniMate on four criteria and their overall average; UniMate scores highest on every axis.}
  \caption{\textbf{User study results on text-conditioned mesh animation.} Mean Likert ratings (1 = very poor, 5 = excellent, with error bars) per criterion and averaged overall; \themodel{} receives the highest rating on all four criteria---text-to-motion agreement, motion plausibility, motion expressiveness, and shape preservation---as well as overall.}
  \label{fig:user-study-results}
\end{figure}

To complement the automatic metrics with a human judgment of perceptual quality, we conducted an anonymous user study on text-driven mesh animations. As illustrated in \cref{fig:user-study-form}, each form presents a single test object together with three side-by-side videos, each generated by a different method from the same input mesh and text prompt; participants then answer the four evaluation questions shown in \cref{fig:user-study-question}. To eliminate ordering and naming bias, the mapping between methods and video labels was anonymized and independently randomized per participant and per example. \Cref{fig:user-study-results} plots the resulting per-criterion ratings.

\section{Additional Results}
\label{app:additional-results}

This section collects additional experimental results that complement the main-paper evaluation: \cref{app:anytop-additional} extends the qualitative AnyTop comparison to further held-out morphologies and motion types, \cref{app:skeleton-free} adds direct visual comparisons against the skeleton-free mesh-animation baselines, \cref{app:qualitative-ablation} visualizes the architecture ablations, \cref{app:curation-ablation} ablates the dataset-curation components, and \cref{app:foot-contact} quantifies foot sliding and the effect of foot-locking post-processing.

\subsection{Additional Comparisons with AnyTop}
\label{app:anytop-additional}

\begin{figure*}[t]
  \centering
  \includegraphics[width= 1.0\textwidth]{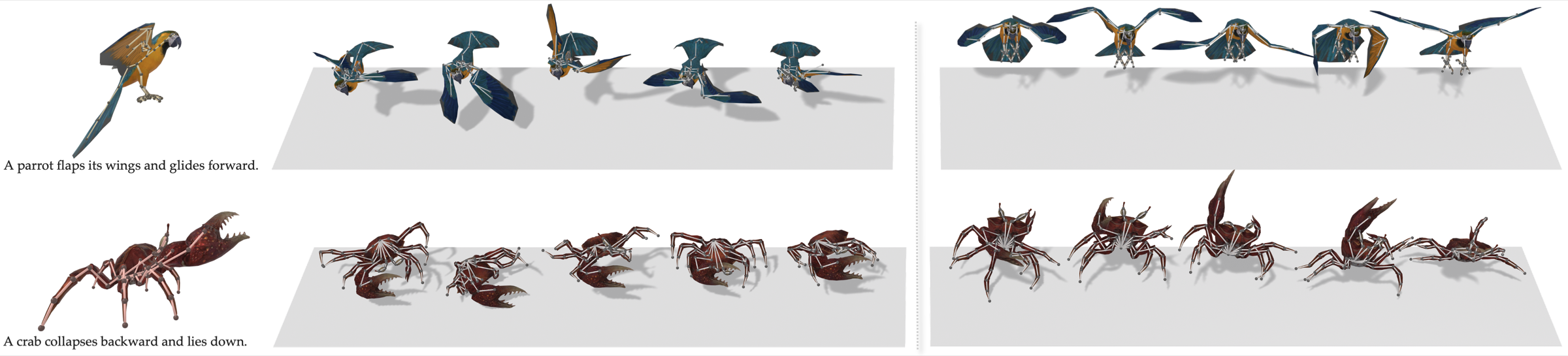}\\[2pt]
  \makebox[ 0.165\textwidth]{}%
  \makebox[ 0.422\textwidth][c]{\small Ours}%
  \makebox[ 0.413\textwidth][c]{\small AnyTop}
  \Description{Two rows comparing UniMate and AnyTop on a gliding parrot and a collapsing crab, with the rest-pose asset and prompt on the left.}
  \caption{\textbf{Additional comparisons with AnyTop on \underline{unseen} Truebones skeletons.} Rest-pose asset and prompt (left), our result (middle), and AnyTop (right), extending the main-paper comparison to an avian glide and a crustacean collapse. Our parrot flaps and glides forward and our crab collapses backward onto its shell as prompted, whereas AnyTop's parrot hovers flapping in place and its crab keeps stepping with raised claws without collapsing.}
  \label{fig:anytop-comparison-additional}
\end{figure*}

\cref{fig:anytop-comparison-additional} extends the AnyTop comparison beyond quadruped and biped locomotion to two further held-out skeletons and motion types from \cref{app:heldout}: an avian rig on \texttt{Parrot2-Land} and a crustacean rig on \texttt{Crab-Die}. The same pattern holds across these morphologies: \themodel{} executes the prompted action on the unseen rig, while AnyTop's outputs remain close to an idle, in-place motion regardless of the prompt.

\subsection{Comparisons with Skeleton-Free Baselines}
\label{app:skeleton-free}

\begin{figure*}[t]
  \centering
  \includegraphics[width= 1.0\textwidth]{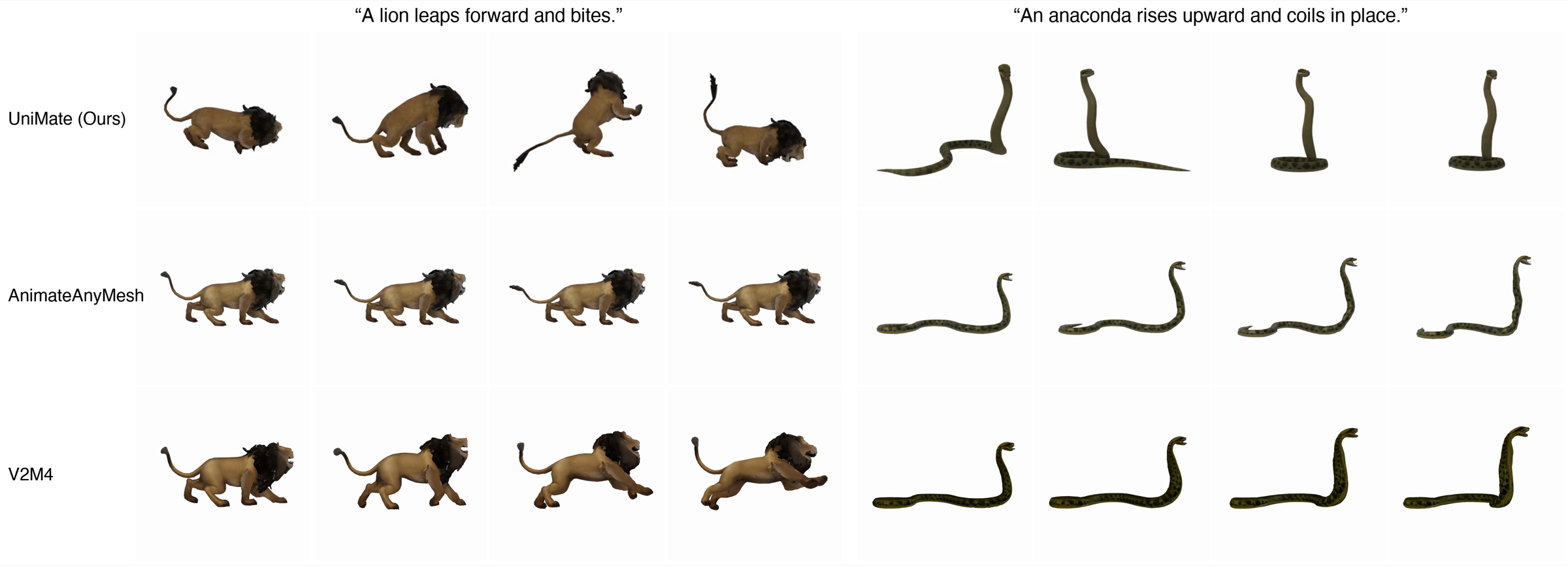}
  \Description{Two blocks of frame sequences comparing UniMate, AnimateAnyMesh, and V2M4 on a leaping lion and an anaconda coiling upward.}
  \caption{\textbf{Qualitative comparison with skeleton-free baselines.} Four evenly spaced frames per clip on two user-study cases, cropped around the subject for visibility. \themodel{} executes the prompted action with dynamic, coherent motion---the lion's leap-and-bite and the anaconda's upward coil; AnimateAnyMesh remains near-static on both cases, while V2M4 distorts the lion's mane geometry and raises only the anaconda's head, never producing the prompted coil.}
  \label{fig:skeleton-free-comparison}
\end{figure*}

\cref{fig:skeleton-free-comparison} contrasts the three methods on two cases from the user-study test set (\cref{app:user-study}). The frame strips visualize the failure modes behind the quantitative gap in the main paper: AnimateAnyMesh's high motion smoothness comes from near-static outputs with little articulated motion, and V2M4, driven by a separately generated video, both deviates from the prompt and accumulates mesh distortion over time. \themodel{} produces prompt-faithful, kinematically coherent motion on both rigs; the project page shows the full clips.

\subsection{Qualitative Ablations}
\label{app:qualitative-ablation}

\begin{figure}[t]
  \centering
  \includegraphics[width=\linewidth]{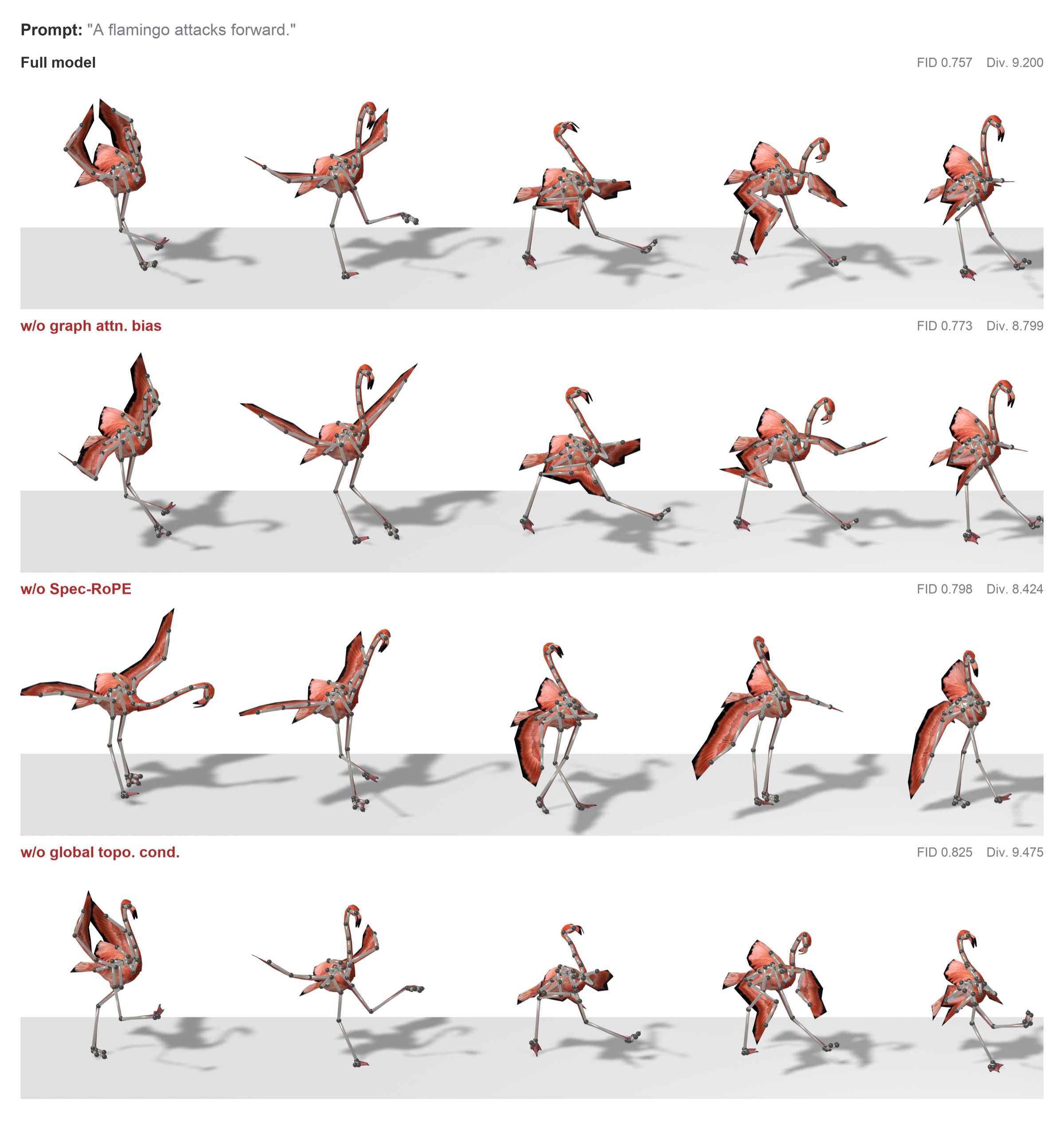}
  \Description{Four rows of flamingo motion frames comparing the full model against three ablated variants, each row annotated with its FID and diversity.}
  \caption{\textbf{Qualitative ablation of the topology-aware components.} The same rig and prompt rendered from the already-trained ablation models of the main paper, with each variant's FID and diversity inset. The full model executes a clear forward attack; without the graph-aware attention bias the wing--body coordination degrades; without Spec-RoPE the motion loses the prompted action and collapses toward in-place wing flailing; without the global topological conditioner the motion stays dynamic but becomes unstable, with jittery, poorly grounded poses.}
  \label{fig:ablation-qualitative}
\end{figure}

\cref{fig:ablation-qualitative} complements the quantitative ablation table of the main paper by rendering the same held-out rig and prompt with each ablated variant. The visual differences track the numbers: the graph-aware attention bias mainly sharpens joint coordination, Spec-RoPE is what carries the prompted action onto an unseen skeleton, and the global topological conditioner stabilizes the motion---removing it raises diversity but visibly degrades stability, which matches its higher diversity score alongside the largest FID increase.

\subsection{Dataset-Curation Ablation}
\label{app:curation-ablation}

\begin{figure}[t]
  \centering
  \includegraphics[width=\linewidth]{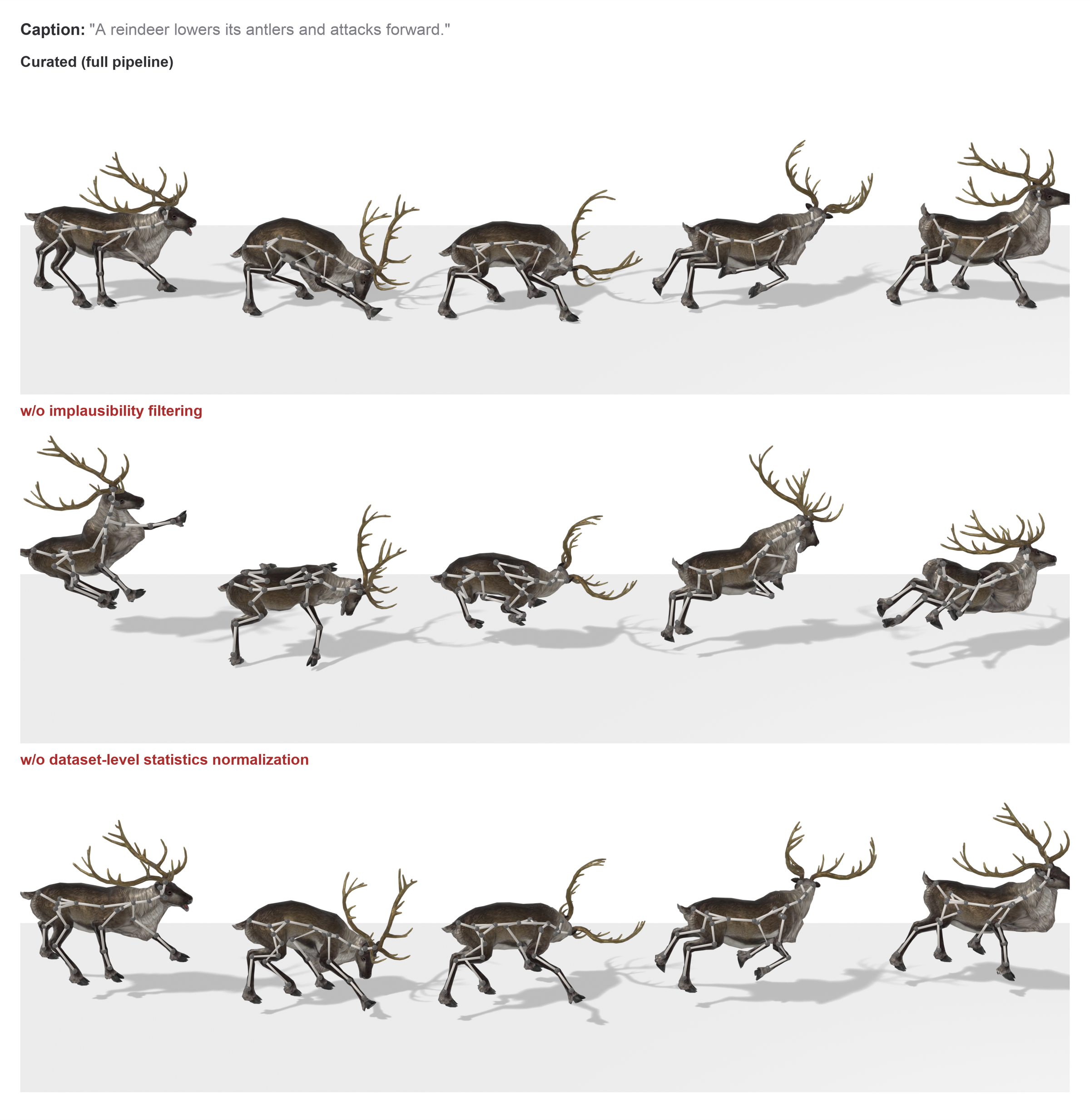}
  \Description{Three rows of reindeer motion frames for the same attack prompt: the full curated pipeline produces a grounded antler attack, the variant without implausibility filtering tumbles into airborne, twisted poses, and the variant without dataset-level statistics normalization drifts and hovers with poor ground contact.}
  \caption{\textbf{Before/after qualitative examples for the curation stages.} The same held-out rig and prompt rendered from models trained with the full curation pipeline (top) and with one stage removed. Without implausibility filtering, the implausible clips that survive in the corpus surface at generation time: the reindeer tumbles through airborne, twisted poses instead of attacking. Without dataset-level statistics normalization, the attack is roughly executed but the motion drifts and hovers with degraded ground contact. The full pipeline performs the prompted antler attack with stable, grounded motion.}
  \label{fig:curation-ablation}
\end{figure}

We ablate the two dataset-curation components that admit a controlled comparison on a fixed dataset. Removing rest-pose rotation rebasing degrades FID from 0.757 to 1.179: expressing joint rotations relative to the rest pose yields a unified kinematic basis across heterogeneous skeletons, and dropping it costs the most motion quality of any curation stage. Disabling the on-the-fly augmentations reduces diversity from 9.200 to 8.265, confirming that they mainly broaden topological coverage rather than per-clip fidelity. The filtering and canonicalization stages change which clips and skeletons enter the dataset, so a direct FID comparison against the unfiltered corpus is not distribution-matched; their per-stage criteria and effects are documented in \cref{sec:dataset} and \cref{alg:uniml3d-data-processing}; as a diagnostic statistic, implausibility filtering discards 8.7\% of the candidate motion clips. \cref{fig:curation-ablation} complements these statistics with before/after qualitative examples: it renders the same held-out rig and prompt from models trained without implausibility filtering (the last filtering stage) and without dataset-level statistics normalization (the last canonicalization stage), alongside the full pipeline. Dropping the filtering stage lets physically implausible training clips leak into the model, which reproduces their erratic, airborne dynamics; dropping the normalization destabilizes ground contact, producing drift and hovering; the full pipeline executes the prompted action with plausible, grounded motion.

\subsection{Foot-Sliding Evaluation and Foot-Locking Post-Processing}
\label{app:foot-contact}

We quantify the foot-sliding artifacts discussed in the limitations section of the main paper, and measure how much of this sliding a rig-level foot-locking post-process removes. Ground contact is only well defined for morphologies with a support pattern, so we evaluate on 20 held-out legged skeletons (10 bipedal and 10 quadrupedal) drawn from the Truebones and Mixamo test splits; serpentine, marine, in-flight, and articulated-object rigs are excluded. For each rig, the set of \emph{contact joints} $\mathcal{C}$ is selected automatically from the standardized joint vocabulary of \cref{app:joint-name-standardization}: the distal-most joint of every limb chain whose canonical name contains a ground-contact term (\emph{Foot}, \emph{Toe}, \emph{Ball}, \emph{Paw}, \emph{Hoof}, or \emph{Pastern}). No per-rig manual annotation is involved.

\paragraph{Normalization.}
Because the evaluated skeletons differ widely in scale and proportion, all lengths are reported in units of the character's rest-pose \emph{root height} $L$, the vertical distance from the skeleton root to the ground plane; the ground plane itself is the horizontal plane through the lowest rest-pose joint. For a contact joint $j$ at frame $t$, let $h_{j,t}$ denote its signed height above the ground plane (negative below it) and $d_{j,t}$ the horizontal (ground-parallel) displacement of the joint between frames $t{-}1$ and $t$, both in units of $L$, with all sequences sampled at 30\,FPS. A joint is \emph{near the ground} when $h_{j,t} < \delta_h$ with $\delta_h = 0.05$, and a near-ground frame counts as \emph{skating} when $d_{j,t} > \delta_s$ with $\delta_s = 0.025$; these thresholds correspond to the 5\,cm and 2.5\,cm used at human scale by~\citet{gmd}, where the hip height is roughly one meter.

\paragraph{Metrics.}
We report two sliding measures, each averaged over all generated sequences of the evaluation set (lower is better). The \textbf{skating ratio (Skate)} follows~\citet{gmd}: the fraction of near-ground frames of contact joints that are skating, i.e.\ $\Pr\left[d_{j,t} > \delta_s \mid h_{j,t} < \delta_h\right]$, which captures how often planted limbs drift. The \textbf{sliding distance (Slide)} is the height-weighted horizontal drift of~\citet{mann}, $s_{j,t} = d_{j,t}\,\bigl(2 - 2^{\,h_{j,t}/\delta_h}\bigr)$, averaged over near-ground frames and reported in units of $10^{-2}L$; unlike the binary ratio, it measures how far the limbs slide.

\paragraph{Foot-locking post-processing.}
Where contact is well defined, foot locking applies zero-shot to the predicted rig without retraining. Per contact joint, frames with $h_{j,t} < \delta_h$ and $d_{j,t} < \delta_s$ are marked as contact, the binary signal is median-filtered with a five-frame window, and segments shorter than three frames are discarded. Each remaining segment is assigned an anchor: the mean horizontal position of the joint over the segment, projected onto the ground plane. The joint is then pinned to its anchor by damped least-squares inverse kinematics over its limb chain (the contact joint and up to three parent joints), leaving the root trajectory and all other chains untouched, and the correction is blended in and out linearly over five frames at each segment boundary to avoid pops.

\begin{table}[!h]
  \centering
  \caption{\textbf{Foot-sliding evaluation on held-out legged skeletons.} Skate is the skating ratio and Slide the height-weighted sliding distance (\cref{app:foot-contact}), the latter in units of $10^{-2}L$ for a rest-pose root height $L$. FL denotes the foot-locking post-process.}
  \label{tab:foot-contact}
  \fontsize{6.5}{8}\selectfont
  \setlength{\tabcolsep}{12pt}
  \begin{tabular}{@{}lcc@{}}
    \toprule
    \textbf{Method} & \textbf{Skate}$\downarrow$ & \textbf{Slide}$\downarrow$ \\
    \midrule
    \themodel\ (Ours) & 0.107 & 0.542 \\
    \themodel\ + FL & \textbf{0.023} & \textbf{0.191} \\
    \bottomrule
  \end{tabular}
\end{table}

\cref{tab:foot-contact} reports the results. Foot locking reduces the skating ratio from 0.107 to 0.023 and the sliding distance from 0.542 to 0.191. Within detected contact segments the correction eliminates sliding by construction; the residual stems from near-ground frames outside detected segments, the blended segment boundaries, and brief touch-downs discarded by the minimum-length filter.

\section{Limitations and Future Work}
\label{app:limitations}
The main paper summarizes the principal contact, scope, and data-coverage limitations of \themodel; here we provide a more detailed discussion and outline additional directions for controllability.

\paragraph{Supervision is bounded by 4D animation data.}
\themodel{} is trained end-to-end on \thedata, a curated corpus of paired skeletal motions and text prompts assembled from artist-authored 4D animation sources. Although \thedata{} is, to our knowledge, the largest cross-species rigged-motion dataset assembled to date, it remains orders of magnitude smaller than the visual corpora available to image and video foundation models, and its category distribution is heavily long-tailed: bipedal humanoids dominate, while serpentine, marine, and insectoid morphologies are sparsely populated (\cref{app:data-sources}). Even after the square-root-balanced sampler and the kinematics-preserving online augmentations described in \cref{app:augmentation}, rare motions (extreme gymnastic skills, fine manipulation, complex social interactions) and underrepresented species (large invertebrates, exotic aquatic locomotion) are visibly harder for the model to cover than well-represented humanoid locomotion. A natural next step is to distill Internet-scale video or video-generative priors~\cite{dimo,vips} into our unified framework---for example, by using video-derived motion fields as auxiliary supervision, or by aligning \themodel's latent dynamics with a pretrained video diffusion model. This would let cross-species generalization scale with passive video data rather than with the cost of new 4D capture, while keeping the topology-aware backbone of \themodel{} intact.
Agentic content-generation systems offer a complementary data-side remedy: Articraft~\cite{articraft}, for example, uses an LLM to synthesize diverse 3D assets at scale, and pairing such generated assets with scripted, simulated, or distilled motion could help densify the rare topology and motion regions that current 4D corpora leave uncovered.

\paragraph{Conditioning is restricted to skeletons and text.}
The current conditioning interface accepts only a rigged skeleton and a natural-language prompt. This is sufficient for the text-to-animation setting we evaluate, but it leaves out several input modalities that artists and downstream systems regularly want to drive animation with: monocular video (motion capture from a single camera), exemplar motion clips (``animate this rig in the style of that clip''), shape-only inputs (a mesh without an authored rig), and partial skeletal demonstrations (e.g., a keyframed root trajectory, a hand pose, or a footstep schedule). Each of these is straightforward to express as an additional conditioning stream consumed by the same AdaLN-Zero injection pathway used for text and topology (\cref{app:conditioning}); the substantive work lies in collecting paired data and designing modality-specific encoders that share \themodel's topology-aware token layout. Adding these channels would move \themodel{} closer to a general motion-capture and animation system for arbitrary 3D assets, in which the same backbone covers text-driven synthesis, video-based mocap, and example-based retargeting under one model.

\paragraph{No native channel for fine-grained spatial motion control.}
Text is a powerful but coarse controller: a prompt can specify ``walk forward'' or ``perform a spin kick,'' but cannot precisely place a foot on a target, route a hand along a desired curve, or impose contact constraints with the environment. \themodel{} currently exposes no first-class mechanism for such spatial directives, which is a real limitation for production use cases where animators need frame-level control over a few key joints. Two complementary directions can address this without retraining the backbone. First, sampling-time guidance methods---for example, the projection-based flow guidance of \citet{projflow}---let users specify per-joint spatial targets and have the ODE solver project each integration step onto the constraint manifold, trading a small amount of fidelity for hard control. Second, in-context learning over partial trajectory tokens, in the spirit of \citet{umo}, would let artists ``paint'' a desired motion onto a subset of joints and frames and have \themodel{} infill the remainder while respecting the topology-aware priors learned during training. Combining these mechanisms with the existing text and topology conditions would strengthen \themodel's value as an animation foundation model that allows artists to both describe and directly control desired motions.

  \putbib[references]
\end{bibunit}

\end{document}